\documentclass[10pt,twocolumn,letterpaper]{article}

\usepackage{cvpr}              % To produce the CAMERA-READY version
\definecolor{cvprblue}{rgb}{0.21,0.49,0.74}
\usepackage[pagebackref,breaklinks,colorlinks,allcolors=cvprblue]{hyperref}
\usepackage{subcaption}
\usepackage{tabularx}
\usepackage{booktabs} % For better table lines
\usepackage{multirow}
\usepackage{multicol}
\usepackage{adjustbox}
\usepackage{rotating}    % For rotating tables

\newcommand\blfootnote[1]{%
  \begingroup
  \renewcommand\thefootnote{}\footnote{#1}%
  \addtocounter{footnote}{-1}%
  \endgroup
}

\def\paperID{10426} % *** Enter the Paper ID here
\def\confName{CVPR}
\def\confYear{2025}

\title{GenDeg: Diffusion-based Degradation Synthesis for Generalizable All-In-One Image Restoration}

\author{
Sudarshan Rajagopalan, Nithin Gopalakrishnan Nair, Jay N. Paranjape and Vishal M. Patel\\
Johns Hopkins University
\\
{\tt\small \{sambasa2, ngopala2, jparanj1, vpatel36\}@jhu.edu
}
}

\begin{document}
\maketitle

\begin{abstract}
    Deep learning–based models for All-In-One image Restoration (AIOR) have achieved significant advancements in recent years. However, their practical applicability is limited by poor generalization to samples outside the training distribution. This limitation arises primarily from insufficient diversity in degradation variations and scenes within existing datasets, resulting in inadequate representations of real-world scenarios. Additionally, capturing large-scale real-world paired data for degradations such as haze, low-light, and raindrops is often cumbersome and sometimes infeasible. In this paper, we leverage the generative capabilities of latent diffusion models to synthesize high-quality degraded images from their clean counterparts. Specifically, we introduce GenDeg, a degradation and intensity-aware conditional diffusion model, capable of producing diverse degradation patterns on clean images. Using GenDeg, we synthesize over $550$k samples across six degradation types: haze, rain, snow, motion blur, low-light, and raindrops. These generated samples are integrated with existing datasets to form the GenDS dataset, comprising over $750$k samples. Our experiments reveal that image restoration models trained on GenDS dataset exhibit significant improvements in out-of-distribution performance as compared to when trained solely on existing datasets. Furthermore, we provide comprehensive analyses on implications of diffusion model-based synthetic degradations for AIOR. 
    \blfootnote{Project Page: \href{https://sudraj2002.github.io/gendegpage/}{\textcolor{blue}{https://sudraj2002.github.io/gendegpage/}}}
\end{abstract}
\section{Introduction}
\label{sec:intro}

\begin{figure}[t]
    \centering
    \setlength{\tabcolsep}{1pt}
    \begin{tabular}{cc}
         \includegraphics[height=4.1cm, width=4.1cm]{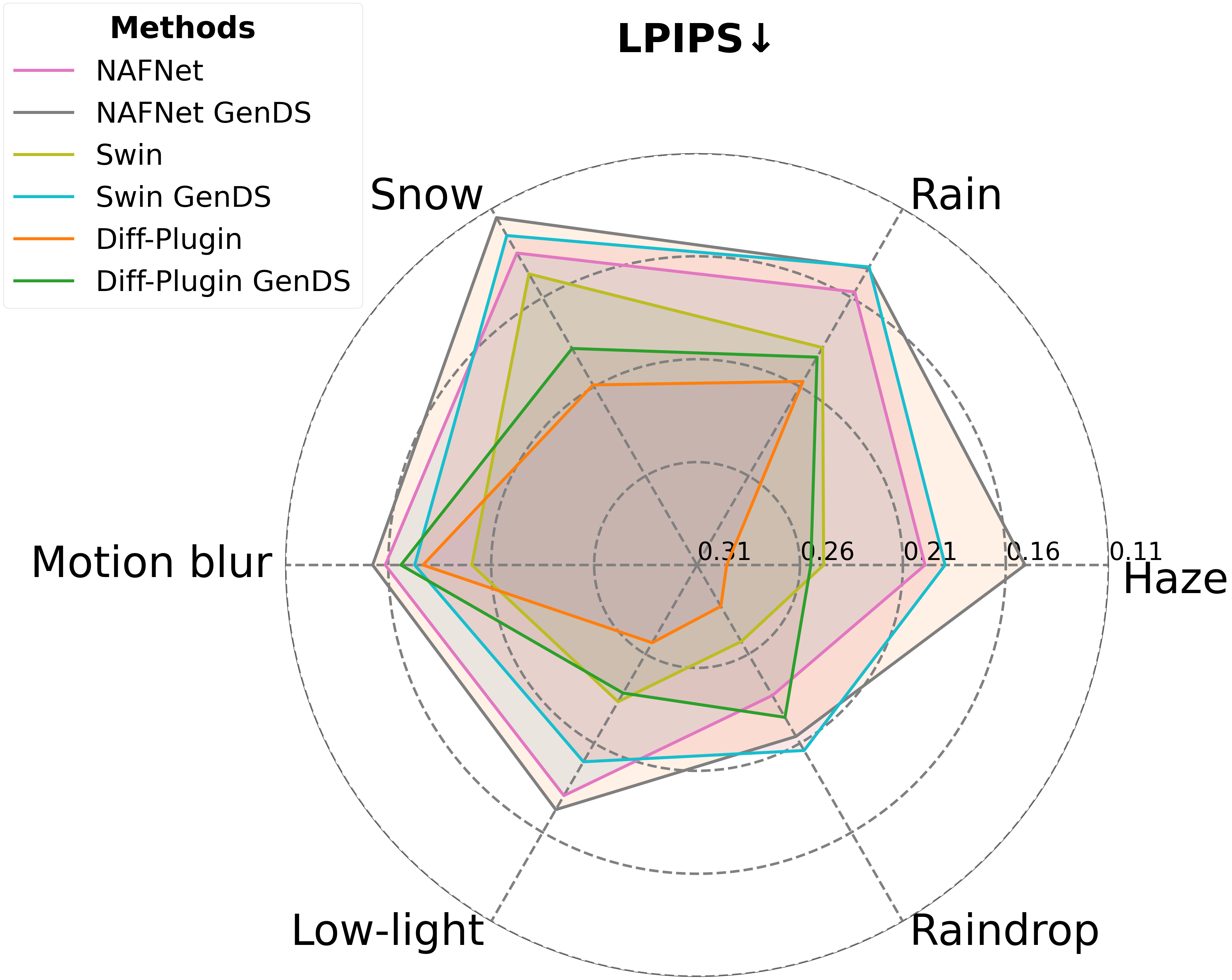}&\includegraphics[height=4.1cm, width=4.1cm]{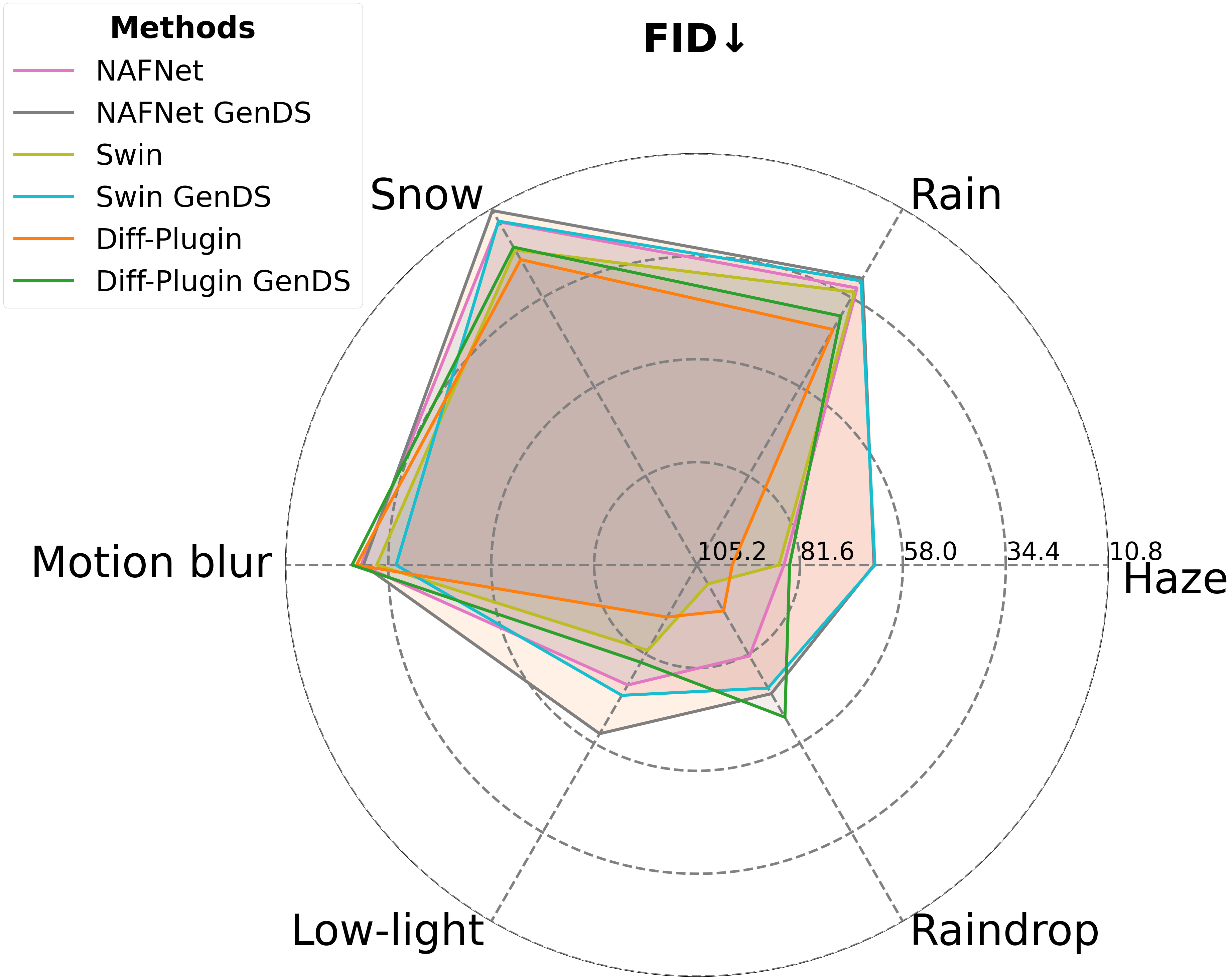}  \\ 
    \end{tabular}
    % \vskip-5pt
    \caption{Out-of-distribution performance of three image restoration models when trained solely using existing datasets and our proposed GenDS dataset. Significant improvements can be observed across all degradations. Metric values reduce outward.}
    \label{fig: main}
    % \vskip-8pt
\end{figure}

Image restoration is a well-studied computer vision problem that aims to reverse the effects of image corruptions or artifacts. It is important for numerous applications, including autonomous driving, imaging and surveillance. Early approaches focused on handling specific degradations such as haze~\cite{early1,haze1}, rain~\cite{early3,rain1,rainnew1,rainnew2}, snow~\cite{snow1,snow2}, blur~\cite{gopro,hide} etc. More recent methods such as Restormer~\cite{restormer}, MPRNet~\cite{mprnet} and SwinIR~\cite{swinir} proposed architectures capable of addressing any single restoration task. However, these approaches are limited to addressing one type of degradation at a time, making them inefficient for scenarios involving multiple types of corruptions.
%due to the need for separate sets of weights for each task.

\begin{figure}[t]
    \centering
    \includegraphics[height=0.72\linewidth, width=1\linewidth]{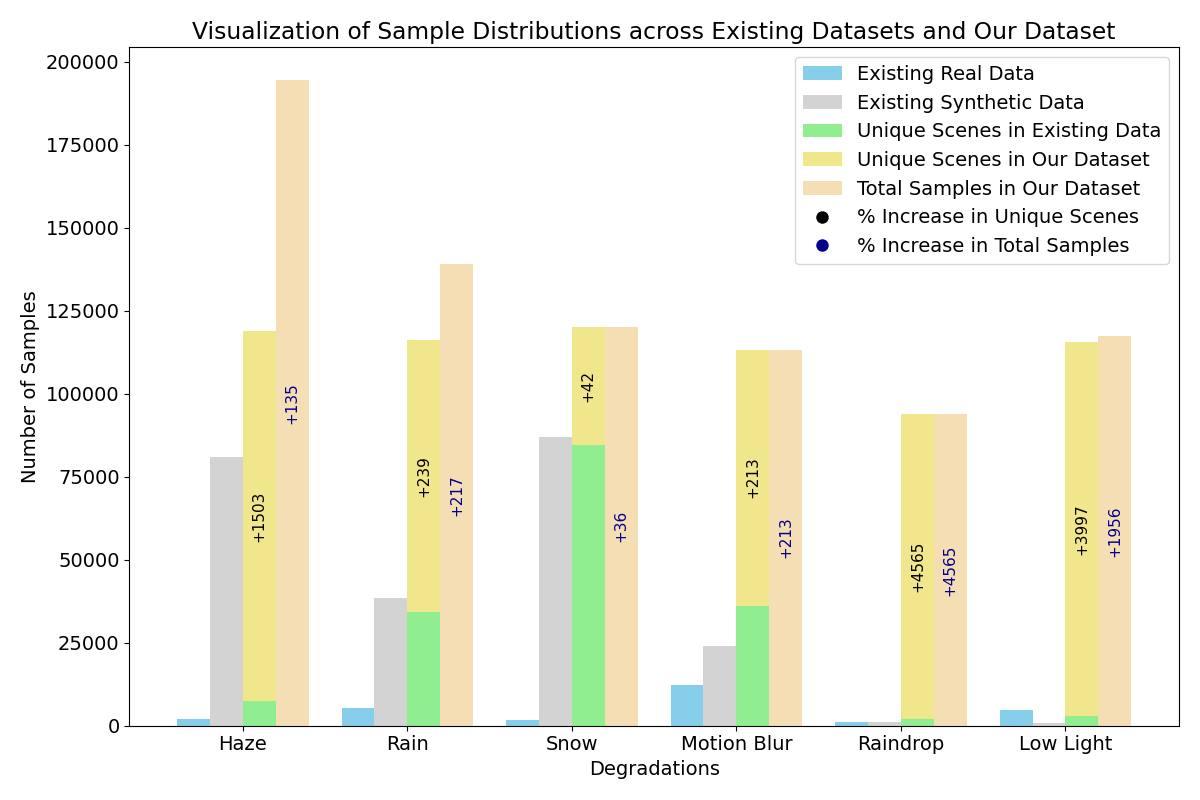}
    % {figs/datasets/bar_plot_even_bigger.png}
    \vskip-10pt
    \caption{Analysis of real and synthetic image restoration datasets for various degradations. Existing datasets are small and less diverse, especially for haze, low-light, and raindrop. Our diffusion-generated synthetic data substantially increases the number of samples as well as scene diversity.}
    \label{fig: datastats}
    % \vskip-8pt
\end{figure}

All-In-One Restoration (AIOR) methods overcome this limitation by employing a single model capable of handling multiple types of degradations. Recent approaches include PromptIR~\cite{promptir}, DA-CLIP~\cite{daclip}, DiffUIR~\cite{diffuir}, Diff-Plugin~\cite{diffplugin}, InstructIR~\cite{instructir} and AutoDIR~\cite{autodir}. Most AIOR methods are trained using a single dataset for each restoration task such as RESIDE~\cite{reside} for dehazing, Snow100k~\cite{snow100k} for desnowing, Rain13K~\cite{mprnet} for deraining and GoPro~\cite{gopro} dataset for motion deblurring. Although these approaches perform well on degradations from these dataset distributions, they often exhibit poor generalization when confronted with new scenes or out-of-distribution (OoD) degradation patterns, which is very common in real-world scenarios. Recent studies have discussed the problem of generalization in image restoration models in great depth~\cite{preliminarygenir, surveygenir}. We hypothesize that the limited generalization of these models is mainly due to two reasons:
%This generalization challenge limits AIOR networks in handling real-world degradations, thereby restricting their practical applicability.

\begin{enumerate}
    \item \textbf{Lack of large datasets with real degradations under diverse scenes.} We consider the degradations haze, snow, rain, raindrop, motion blur and low-light. Fig.~\ref{fig: datastats} shows the number of synthetic and real images for each degradation from its existing publicly available datasets, along with the number of unique scenes. To the best of our knowledge, we have included most of the existing datasets. Firstly, the figure shows that existing restoration datasets are significantly smaller than those used to train generalizable models for other low-level vision tasks, such as SAM~\cite{sam} for segmentation and Depth-Anything~\cite{depthanything} for depth estimation ($>1.5$M samples). This limited dataset size hinders the generalization of models to diverse real-world scenarios. Secondly, degradations such as haze, raindrop, low-light and snow have very few real images compared to synthetic ones due to challenges in capturing real images under these conditions. For instance, haze is an atmospheric phenomenon which is difficult to simulate in real scenarios. Conversely, motion blur and rain have more real-world examples as they can be generated from existing videos~\cite{gopro,hide,rain1, realrain1k, lhprain}. Thirdly, degradations such as haze, raindrop and low-light have very limited scene diversity which can further limit the generalization of models. Finally, the number of samples across different degradations is highly imbalanced.

    \item \textbf{Lack of variety in degradation patterns within datasets.} Previously, we analyzed the distribution of samples for each degradation. Examining individual datasets, especially synthetic ones, reveals that they contain degradations generated using only a particular model. For instance, the images in the RESIDE~\cite{reside} dataset are generated by the atmospheric haze model~\cite{hazemodel1} with specific parameters. Consequently, training a network on such a dataset can tailor it to work only for the degradation patterns of that dataset, limiting its generalization capability for real-world dehazing. 
\end{enumerate}

Thus, AIOR methods often overfit to training distributions, limiting generalization to real-world samples. To address this, we aim to develop robust AIOR models capable of generalizing to OoD restoration. 
We define OoD samples as those from test sets, whose corresponding training sets are not used. Within-distribution test sets refer to test splits of the datasets used for training (also referred to as existing training data). To support these definitions, we compute the Wasserstein Distance (WaD) between DA-CLIP~\cite{daclip}'s degradation feature distributions for OoD test sets and the existing training datasets for each degradation (see Table~\ref{tab: wad}). The mean WaD between within-distribution test sets and their training sets is $0.084$, which is much lower than the WaD for OoD test sets, validating our definition. Achieving good OoD performance requires large-scale training data with diverse degradations. Since collecting real-world data for all the degradations is infeasible, we propose a novel degradation generation framework that leverages the generative capabilities of Latent Diffusion Models (LDMs)~\cite{stablediff} to synthesize diverse degradations. Specifically, we introduce GenDeg, a Stable Diffusion-based model conditioned on text prompts, clean images and degradation intensity to produce images under various degradations.
%Specifically, we define OoD samples as those from test sets, whose corresponding training sets are not used during training. Within-distribution test sets refer to test splits of the datasets used for training (also referred to as existing training data). To support these definitions, we compute the Wasserstein Distance (WaD) between DA-CLIP~\cite{daclip}'s degradation feature distributions for OoD test sets, within-distribution test sets and the existing training datasets for each degradation. The mean WaD between within-distribution test sets and their corresponding training sets is $0.084$, which is much lower than WaD between OoD and existing training sets (see Table~\ref{tab: wad}), validating the distribution shift.

\begin{table}[t]
    \centering
    \small
    \caption{Wasserstein distance between the degradation feature distributions of DA-CLIP~\cite{daclip} for existing training data and OoD test sets.}
    \vskip-8pt
    \setlength{\tabcolsep}{2pt} % Reduce column separation
    \begin{tabular}{lcccccc} % Use fixed-width table
    \toprule
        \textbf{Distance} & 
        \textbf{Haze} & 
        \textbf{Rain} & 
        \textbf{Snow} & 
        \textbf{Motion} & 
        \textbf{Low-light} & 
        \textbf{Raindrop} \\
        \midrule
        
        Wasserstein & 
        0.212& 
         0.287& 
         0.289& 
         0.173& 
         0.349& 
         0.177\\
        \bottomrule
    \end{tabular}
    \label{tab: wad}
    \vskip-8pt
\end{table}

\begin{figure}[t]
    \centering
    \includegraphics[height=0.6\linewidth, width=1\linewidth]    {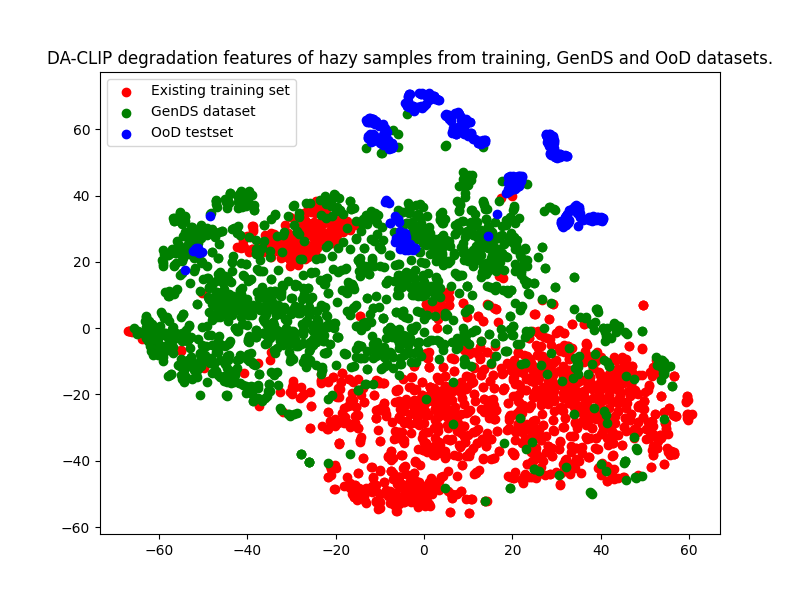}
    % {figs/analysis/haze_daclip_new.png}

    \vskip-10pt
    \caption{t-SNE visualization of degradation features obtained for hazy samples from existing training data, GenDS dataset and OoD test sets. The features were obtained using DA-CLIP~\cite{daclip}.}
    \label{fig: tsne}
    \vskip-8pt
\end{figure}

%Latent diffusion models (LDMs)~\cite{stablediff} excel at generating diverse high-quality images. We leverage their generative capability for synthesizing diverse degradations. Specifically, we propose GenDeg, a Stable Diffusion-based model conditioned on text prompts, clean images and degradation intensity to produce images under different degradations. 
We train GenDeg by combining multiple existing datasets for each degradation type to ensure that it does not heavily rely on a specific degradation pattern or physical model. Thus, it can produce both synthetic and realistic degradations, thereby enriching the diversity of degradation patterns in the generated data. Furthermore, GenDeg offers fine-grained control over the intensity and spatial variations of generated degradations by conditioning on the mean ($\mu$) and standard deviation ($\sigma$) of the degradation map. Using GenDeg, we synthesize over $550$k degraded images from roughly $120$k clean images. We augment existing restoration datasets (used to train GenDeg) with our generated images to create a dataset, GenDS, with over $750$k paired images under haze, rain, snow, motion blur, low-light and raindrop degradations. GenDS provides a significant boost in scene diversity and number of samples (see Fig.~\ref{fig: datastats}), resulting in substantial improvements in OoD performance of AIOR models (see Fig.~\ref{fig: main}). Fig.~\ref{fig: tsne} illustrates that the diverse degradations in the GenDS dataset help bridge the domain gap between existing and OoD datasets (see Sec.~\ref{subsec: analysis} for more details). Additionally, GenDS consists of the same clean images under different degradations that, to the best of our knowledge, is the first such dataset.

Finally, we train five models on the GenDS dataset, namely NAFNet~\cite{nafnet}, PromptIR~\cite{promptir}, a Swin Transformer-based model that we propose, DA-CLIP~\cite{daclip} and Diff-Plugin~\cite{diffplugin}. Our experiments demonstrate that these models achieve significant improvements in their generalization performance when trained on our large-scale dataset.

In summary, our contributions are as follows:
\begin{enumerate}
    \item We propose a novel diffusion model-based degradation generation framework, GenDeg, which is capable of producing diverse degradations on any clean image.

    \item Using GenDeg, we synthesize over $550$k degraded images which when combined with existing datasets forms the comprehensive GenDS dataset comprising approximately $750$k samples across highly diverse scenes. Furthermore, each image in GenDS has multiple degraded versions, making it, to the best of our knowledge, the first restoration dataset of its kind.

    \item Finally, we train restoration models on the GenDS dataset and demonstrate that incorporating our synthetic data significantly improves the out-of-distribution restoration capabilities of these networks.
\end{enumerate}
\section{Related Works}
\label{sec:related}

In this section, we discuss relevant works on all-in-one image restoration and diffusion models for synthetic data. Related works on diffusion models are given in supplementary.

\subsection{All-in-one image restoration}
All-In-One restoration (AIOR) methods employ a single model to address multiple corruptions. Early approaches include All-in-one~\cite{nas}, which used neural architecture search, and Transweather~\cite{transw}, which unified multiple encoders for efficient multi-weather restoration. Airnet and~\cite{tkmc} used contrastive loss to learn degradation representations while PromptIR~\cite{promptir} utilized learnable prompts. Recent approaches have leveraged diffusion models for AIOR. DA-CLIP~\cite{daclip} used degradation information from CLIP~\cite{clip} to guide diffusion-based image restoration. Diff-Plugin~\cite{diffplugin} adopts task plugins to guide a latent diffusion model for restoration. DiffUIR~\cite{diffuir} proposed selective hourglass mapping and AutoDIR~\cite{autodir} developed an automatic approach using vision-language models for degradation detection and restoration. InstructIR~\cite{instructir} utilized text guidance as instructions for AIOR while AWRaCLe~\cite{awracle} used visual in-context learning. Despite these advancements, no prior work (to the best of our knowledge) has explored using diffusion models to generate degradations. Our approach enables the creation of large datasets with realistic degradations to train generalizable AIOR models.

\subsection{Diffusion models for synthetic data}
Diffusion models have shown great potential for synthetic data generation. Methods such as Dreambooth~\cite{dreambooth} and Distribution Matching Distillation~\cite{dmd} focus on improving the generative process using diffusion-generated data, rather than creating paired synthetic data for downstream tasks. Sim2real techniques such as EPE~\cite{epe} translate simulation images into realistic ones but do not synthesize degraded images for restoration.
Recent research has focused on utilizing latent diffusion models for synthetic data generation tailored to specific tasks.~\cite{diffclass1,diffclass2,diffclass3,diffzeroshot1,diffzeroshot2} used diffusion-generated images to improve classification and zero-shot classification performance. However, classification tasks do not require preservation of intricate details in the generated images. Other approaches~\cite{diffusionseg,diffumask,labelfreedom,satsynth} use diffusion-generated data for semantic and aerial segmentation, demonstrating promising directions. Nonetheless, these approaches generate segmentation masks which lack detailed scene content. In contrast, we generate high quality degraded images for image restoration, where preserving scene consistency is crucial for good performance. Our approach effectively addresses these challenges, leading to significant improvements in the generalization of image restoration models trained on our synthetic data.

%demonstrated promising directions. Further,~\cite{labelfreedom,satsynth} showed that augmenting real data with diffusion-generated samples enhances aerial segmentation performance. Nonetheless, these approaches generate only segmentation masks which lack detailed scene content. In contrast, we propose to generate high quality degraded images for image restoration tasks, for which ensuring precise scene consistency is crucial, as discrepancies can degrade restoration performance. Our approach effectively addresses these challenges, leading to significant improvements in the generalization of image restoration models trained with our generated data.
\section{Proposed Method}
\label{sec:proposed}

In this section, we detail GenDeg, our diffusion-based method for generating large-scale synthetic data for image restoration. We also discuss the process of data generation, curation and training of restoration models.

\subsection{Diffusion-based degradation generation}
\label{subsec: diffmodel}

\begin{figure*}
    \centering
    \includegraphics[width=0.9\linewidth]{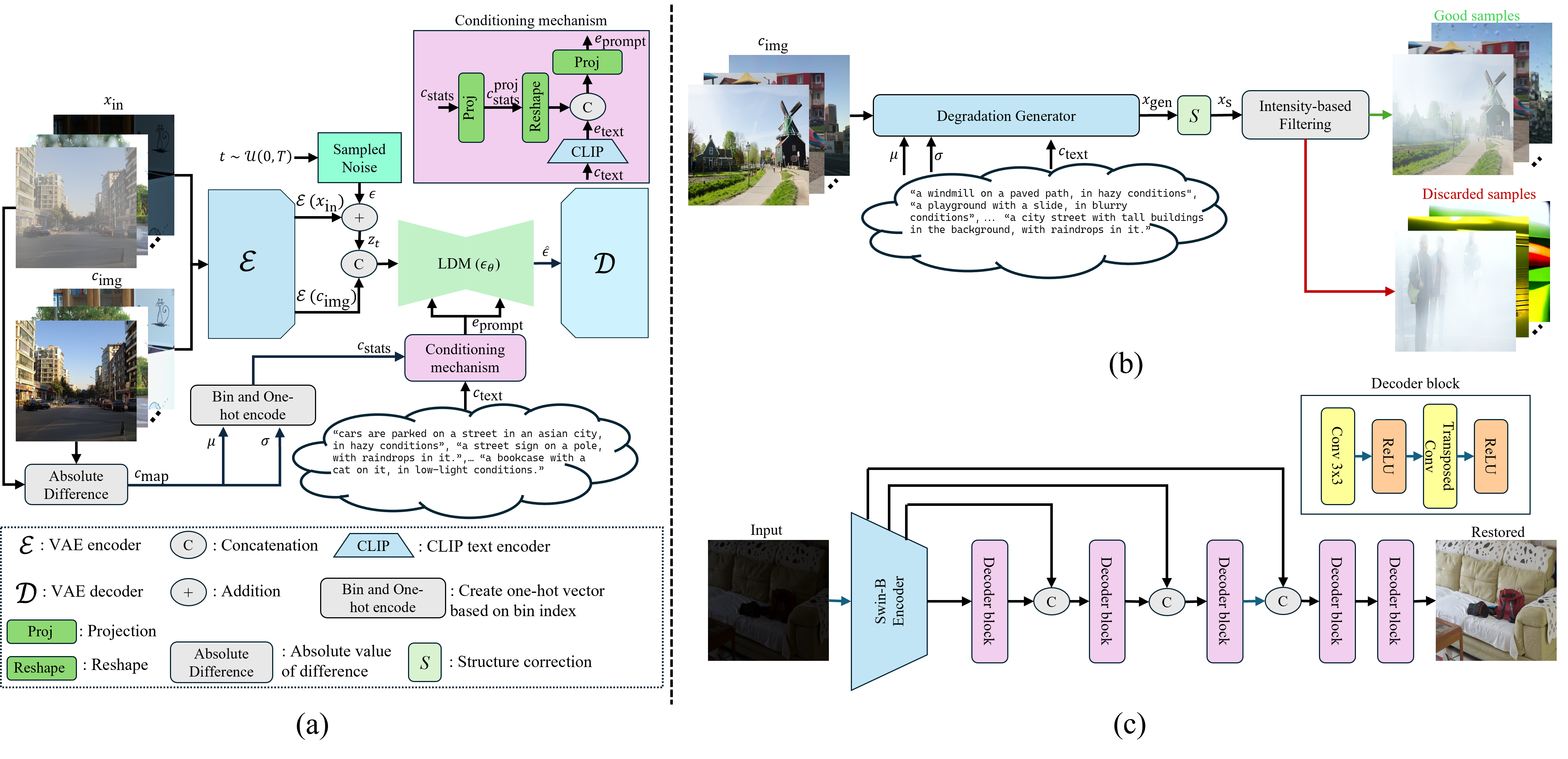}
    \vskip-7pt
    \caption{(a) Illustrates the training stage of the GenDeg model where it is trained to condition on the clean image, text prompt and mean intensity ($\mu$) and variation ($\sigma$) of the degradation pattern. (b) Shows the  inference stage where the model generates a degraded image based on these conditions; and (c) Depicts the architecture of the Swin-transformer-based restoration network.}
    \label{fig: block}
    %\vskip-8pt
\end{figure*}

Our goal is to leverage the generative priors of pre-trained diffusion models to produce diverse degradations on clean images while preserving  scene semantics. We consider the synthesis of six degradations, namely, haze, rain, snow, low-light, motion blur and raindrops. To achieve this, we require a diffusion model that conditions on an input clean image (that needs to be degraded) and a prompt specifying the desired degradation. One popular approach that aligns with our objectives is the InstructPix2pix~\cite{ip2p} model. It is a text-based image editing framework that leverages the latent diffusion model (LDM), Stable Diffusion~\cite{stablediff}, to generate edited images that are consistent with the input image. LDM operates in the latent space of a pre-trained variational auto-encoder~\cite{vae} whose encoder and decoder are denoted by $\mathcal{E}$ and $\mathcal{D}$, respectively. Given an image $x_{\text{in}}$, image condition $c_{\text{img}}$ and text condition $c_{\text{text}}$, the diffusion model ($\epsilon_\theta$) minimizes the following objective during training
\vskip-15pt
\small
\begin{equation}
    L = \mathbb{E}_{\mathcal{E}(x_\text{in}),\,\mathcal{E}(c_\text{img}),\,c_\text{text},\,\epsilon \sim \mathcal{N}(0,1),\,t} \left[ \left\| \epsilon - \epsilon_{\theta}(z_t, t, \mathcal{E}(c_\text{img}), c_\text{text}) \right\|_2^2 \right], 
    \label{eq:diffobj}
\end{equation}
\normalsize
where $z_t$ is the noised version of $\mathcal{E}(x_\text{in})$ at timestep $t$ of the forward diffusion process and $\epsilon$ is the added noise. 

In our adaptation, $x_\text{in}$ is the degraded image from existing paired restoration datasets, $c_\text{img}$ is the corresponding clean image, and $c_\text{text}$ is a text-prompt describing the degradation to be produced. To train the diffusion model, we combine multiple synthetic and real image restoration datasets (see supplementary for details). This approach ensures that the diffusion model learns to produce diverse degradation patterns not specific to any single dataset. 

% We now describe the datasets used for training the diffusion model:

%\begin{enumerate}
%    \item Haze: \textit{I-Haze} dataset-$25$ images for training and $5$ images for testing.
%\end{enumerate}

The text condition $c_\text{text}$ includes high-level scene information along with degradation specifics. We obtain scene descriptions by processing clean images through the BLIP-2~\cite{blip2} captioning model. We then append degradation-specific phrases such as ``, in hazy conditions.'' to these descriptions, forming the final text prompt. Incorporating scene descriptions provides initial guidance to Stable Diffusion during training, helping it generate an image related to $c_\text{img}$. Examples of the text prompts are shown in Fig.~\ref{fig: block}. 

%For our diffusion model, $x_in$ is the degraded image from existing paired restoration datasets, $c_\text{img}$ refers to its corresponding clean image and $c_\text{text}$ is a text prompt conveying high-level information about the scene along with the degradation which needs to be produced. For obtaining such text descriptions, we pass the clean images in restoration datasets through the BLIP 2 image captioning model and obtain a description of the scene. We then append text such as ", in hazy conditions." to these scene descriptions to obtain the final text prompt. We do not directly use the degradation text alone because the scene description acts as initial guidance to Stable Diffusion for producing a scene related to that in $c_\text{img}$. Examples of this data can be found in Fig.~\ref{fig: block}. 

While this method produces degraded images effectively, we observed that using only the degradation type in the prompt causes the diffusion model to generate extreme degradation patterns during inference. For instance, the haze is excessively thick, or the rain is unrealistically heavy or minimal (see supplementary for examples). Such degradations could negatively impact the performance of a restoration model trained on this data, as the patterns differ significantly from typical real-world scenarios. To overcome this limitation, we introduce a conditioning on the level of the degradation, quantified by the mean intensity ($\mu$) and standard deviation ($\sigma$) of the degradation map $c_\text{map}$ defined as $c_\text{map} = |x_\text{in} - c_\text{img}|$. $\mu$ represents mean intensity of degradation in the degraded image while $\sigma$ indicates its spatial distribution across the image. We fuse the conditioning information in the form of $\mu$ and $\sigma$ with the CLIP~\cite{clip} embedding, $e_\text{text}$, of the prompt, $c_\text{text}$, as follows. First, we compute the range, $[a, b]$ of $\mu$ and $\sigma$ for each degradation type from all their respective datasets. We divide this range into $128$ bins and obtain a one-hot encoding for the bins corresponding to particular $\mu$ and $\sigma$ values calculated from $c_\text{map}$ during training. An additional bin is included for null-prompt conditioning~\cite{ip2p}, resulting in vectors of length $129$. 

We then concatenate the one-hot vectors for $\mu$ and $\sigma$ to obtain $c_{\text{stats}} \in \mathbb{R}^{2 \times 129}$. $c_{\text{stats}}$ is then projected to $c^\text{proj}_\text{stats} \in \mathbb{R}^{2 \times 77}$. $c^\text{proj}_\text{stats}$ is then transposed to $\mathbb{R}^{77 \times 2}$ and concatenated with $e_{\text{text}} \in \mathbb{R}^{77 \times 768}$ to obtain a vector of size $\mathbb{R}^{77 \times 770}$. We project this vector back to the CLIP text embedding dimension and obtain $e_{\text{prompt}} \in \mathbb{R}^{77 \times 768}$ to be fed as  conditioning to Stable Diffusion. All projection layers are learnable. This conditioning mechanism ensures that the diffusion model is aware of the degradation level to be added to the clean image, resulting in generated images, $x_{\text{gen}}$, with diverse and realistic degradation patterns. The effect of varying $\mu$ and $\sigma$ is given in Sec.~\ref{subsec: analysis}. Fig.~\ref{fig: block} (a) summarizes the above steps.

Finally, we tackle the challenge of aligning the generated degraded images ($x_{\text{gen}}$) precisely with the input clean images ($c_\text{img}$). The VAE encoding and decoding process in LDMs causes the loss of fine details in the image~\cite{autodir,textdiffuser}. To mitigate this issue, we draw inspiration from AutoDIR~\cite{autodir}, which introduced a Structure Correction Module (SCM) to reverse VAE-induced distortions. In our framework, the SCM, denoted by $S$, corrects $x_{\text{gen}}$ as follows:
\vskip-10pt
\begin{equation}
    x_{S} = x_{\text{gen}} + S([x_{\text{gen}}, c_{\text{img}}])
\end{equation}

The goal of $S$ is to reverse LDM and VAE artifacts while preserving the generated degradation. We train $S$ after the degradation generator has been trained (keeping generator frozen) using a one-step reverse diffusion process:
\vskip-10pt
\begin{equation}
    z_{\text{gen}} = \frac{(z_t - \sqrt{1 - \Bar{\alpha_t}} \cdot \epsilon)}{\sqrt{\Bar{\alpha_t}}}
\end{equation}
Here $\bar{\alpha}_t$ is the cumulative product of the noise schedule up to timestep $t$, and $z_t$ is the noisy latent. We then obtain $x_{\text{gen}} = \mathcal{D}(z_{\text{gen}})$. The loss function for training $S$ is given by
\vskip-10pt
\begin{equation}
    L_S = \sqrt{\Bar{\alpha}_{t-1}} \cdot \sqrt{1 - \Bar{\alpha_t}} \cdot \left\| x_\text{in} - x_{S}\right\|_2^2
\end{equation}
The term $\sqrt{\Bar{\alpha}_{t-1}} \cdot \sqrt{1 - \Bar{\alpha_t}}$ weights the performance of $S$ for each timestep, recognizing that structure correction is easier near the initial timesteps ($t \approx 0$) and quite challenging near the final timesteps ($t \approx T$). This weighting reduces the influence of these extreme cases during training. 

We found that $S$ works well for degradations that possess smooth characteristics such as haze, raindrops and motion blur. However, for degradations such as rain and snow, $S$ blurs out the rain streaks and snowflakes in the generated image. Similarly, for low-light conditions, the SCM can produce blurry outputs due to the low pixel intensities. Thus, for rain, snow and low-light degradations, we omit the usage of $S$. Instead, we pass the clean image through the VAE encoder and decoder to obtain a slightly altered version $\hat{c}_\text{img}$ that is better aligned with the generated image than the original clean image ($c_\text{img}$). Visual results showcasing the effect of $S$ can be found in the supplementary. 

%Future research could focus on developing improved correction modules to handle these specific degradations effectively.

\subsection{Dataset creation}
\label{subsec: synthdataset}

GenDeg enables synthesizing diverse degradations on any clean image. We generate degradations on unique clean images taken from the training datasets of GenDeg. Since we use a large number of training datasets, we obtain approximately $120$k distinct scenes. For each clean image, we produce degradations that were not present in its original training set, resulting in five degradations per image. This strategy supplements existing restoration datasets with our synthetic data, enhancing the generalization capabilities of restoration models when trained on them. 

To generate a particular degradation, we randomly select a dataset associated with that degradation type. We sample $\mu_{\text{gen}}$ from the histogram of $\mu$ values in the selected dataset, which is created by the same binning strategy used during training. Subsequently, we sample $\sigma_{\text{gen}}$ from a similar histogram of $\sigma$ values obtained from images belonging to the sampled $\mu_{\text{gen}}$ bin. This process ensures that the value of $\sigma_{\text{gen}}$ is meaningfully correlated with the chosen $\mu_{\text{gen}}$, resulting in realistic degradation patterns. To further enhance diversity, for every $1$ in $20$ images, we select a random value of $\sigma_\text{gen}$ (within acceptable limits) for a chosen $\mu_\text{gen}$. The clean image is then degraded using the chosen $\mu_\text{gen}$ and $\sigma_\text{gen}$ values. After generation, we filter images based on the mean value of the generated degradation map to discard poor quality images (see Fig.~\ref{fig: block} (b)). The filtering thresholds are empirically determined for each degradation type (given in supplementary). In total, after filtering, we create approximately $550$k degraded images which are combined with samples from existing datasets to obtain the GenDS dataset.

\newcolumntype{C}{>{\centering\arraybackslash}X}

\begin{table*}[t]
    \centering
    \caption{Quantitative comparisons of NAFNet~\cite{nafnet}, PromptIR~\cite{promptir}, Swin transformer, DA-CLIP~\cite{daclip} and Diff-Plugin~\cite{diffplugin} models using LPIPS and FID metrics (lower is better), trained with and without our GenDS dataset. Performance is evaluated on OoD test sets. The table also includes the performance of existing state-of-the-art (SOTA) approaches. Training with the GenDS dataset significantly enhances OoD performance. (R) indicates real images and (S) indicates synthetic images. Diff-Plugin\textsuperscript{\#} is the publicly available pre-trained model.}
    \vskip-8pt
    \small
    \label{tab: quant_ood}
    \begin{tabularx}{\textwidth}{l *{9}{C}}  % 'l' for Method, 'C' for others
        \toprule
        \textbf{Method} & 
        \textbf{REVIDE \cite{revide}} & 
        \textbf{O-Haze \cite{ohaze} } & 
        \textbf{RainDS \cite{rainds} } & 
        \textbf{LHP \cite{lhprain} } & 
        \textbf{RSVD \cite{rsvd} } & 
        \textbf{GoPro \cite{gopro} } & 
        \textbf{LOLv1 \cite{lolv1} } & 
        \textbf{SICE \cite{sice} } & 
        \textbf{RainDS \cite{rainds} } \\
        \midrule

        \textbf{Degradation Type} & 
        Haze (R) & 
        Haze (R) & 
        Rain (S)& 
        Rain (R)& 
        Snow (S)& 
        Motion Blur (R)& 
        Low-light (R)& 
        Low-light (R)& 
        Raindrop (R)\\
        \midrule
        
        DiffUIR & 
        0.268/58.5 & 
        0.334/147.2 & 
        0.088/31.2 & 
        0.187/26.5 & 
        0.176/26.1 & 
        0.144/25.2 & 
        0.148/65.1 & 
        0.442/102.9 & 
        - \\
        
        Diff-Plugin\textsuperscript{\#} & 
        0.281/72.9 & 
        0.377/164.7 & 
        0.194/45.0 & 
        0.178/30.2 & 
        0.207/22.5 & 
        0.217/32.8 & 
        0.195/70.5 & 
        0.233/69.2 & 
        - \\
        
        InstructIR & 
        0.313/65.4 & 
        0.341/154.5 & 
        0.117/29.2 & 
        0.139/21.2 & 
        - & 
        0.146/21.1 & 
        0.132/57.3 & 
        0.234/65.2 & 
        - \\
        
        AutoDIR & 
        0.247/57.9 & 
        0.315/144.1 & 
        0.105/30.6 & 
        0.181/27.0 & 
        - & 
        0.157/22.2 & 
        0.116/43.7 & 
        0.249/74.0 & 
        0.157/52.4 \\
        \midrule
        
        PromptIR & 
        0.262/62.0 & 
        0.333/150.9 & 
        0.111/49.3 & 
        0.186/29.3 & 
        0.128/15.8 & 
        0.186/32.9 & 
        0.258/111.8 & 
        0.391/99.3 & 
        0.208/106.8 \\
        
        PromptIR GenDS & 
        0.212/56.0 & 
        0.160/89.0 & 
        0.096/34.4 & 
        0.182/28.1 & 
        0.119/13.9 & 
        0.191/31.9 & 
        0.178/87.9 & 
        0.375/90.7 & 
        0.182/79.8 \\
        
        Swin & 
        0.242/62.9 & 
        0.254/109.9 & 
        0.182/38.8 & 
        0.189/27.1 & 
        0.143/21.7 & 
        0.198/31.7 & 
        0.241/112.0 & 
        0.293/88.3 & 
        0.232/82.6 \\
        
        Swin GenDS & 
        0.209/54.3 & 
        0.165/74.6 & 
        0.116/35.8 & 
        0.162/24.1 & 
        0.121/14.1 & 
        0.170/36.2 & 
        0.167/73.1 & 
        0.241/72.1 & 
        0.197/70.7 \\
        
        NAFNet & 
        0.211/71.3 & 
        0.183/99.2 & 
        0.107/34.4 & 
        0.200/29.3 & 
        0.131/14.3 & 
        0.155/28.2 & 
        0.167/78.8 & 
        0.304/83.5 & 
        0.178/73.4 \\
        
        NAFNet GenDS & 
        0.151/52.5 & 
        0.143/76.7 & 
        0.100/31.5 & 
        0.180/27.1 & 
        0.110/11.3 & 
        0.149/28.7 & 
        0.147/63.7 & 
        0.278/78.5 & 
        0.170/60.5 \\

        DA-CLIP  & 
        0.300/59.7 & 
        0.369/154.6 & 
        0.153/49.1 & 
        0.194/39.7 & 
        0.177/39.0 & 
        0.187/39.1 & 
        0.332/100.8 & 
        0.331/91.7 & 
        0.254/85.1 \\

        DA-CLIP GenDS  & 
        0.215/51.4 & 
        0.251/113.7 & 
        0.140/45.9 & 
        0.178/35.9 & 
        0.148/30.7 & 
        0.146/33.5 & 
        0.250/81.6 & 
        0.285/76.1 & 
        0.246/78.9 \\

        Diff-Plugin  & 
        0.273/63.9 & 
        0.320/130.3 & 
        0.197/49.3 & 
        0.214/36.4 & 
        0.207/24.3 & 
        0.174/27.1 & 
        0.276/103.3 & 
        0.256/79.5 & 
        0.287/93.0 \\

        Diff-Plugin GenDS & 
        0.230/57.1 & 
        0.278/110.8 & 
        0.186/45.2 & 
        0.197/33.3 & 
        0.186/21.0 & 
        0.163/26.0 & 
        0.252/90.9 & 
        0.223/68.0 & 
        0.223/64.9 \\
        
        \bottomrule
    \end{tabularx}
    \vskip-8pt
\end{table*}

\subsection{Training image restoration models}
\label{subsec: genirmodel}

%Current restoration models (aside from recent LDM based architectures) do not leverage a pre-trained initializations that are capable of providing robust foundational features. 

Transformer-based architectures have demonstrated enormous potential in learning generalizable image features. However, the usage of pre-trained transformers for image restoration remains limited. We hypothesize that transformer encoders pre-trained on large datasets such as ImageNet~\cite{imagenet} can serve as effective feature encoders for improving generalization in restoration tasks. Hence, we choose a pre-trained Swin transformer encoder~\cite{simmim} for extracting generalizable features from degraded images. We specifically choose the Swin transformer~\cite{swin} over the standard Vision Transformer (ViT) as it provides hierarchical features at multiple resolutions, which is crucial for preserving fine details in restored images. To reconstruct the restored image from the features extracted by the Swin transformer, we employ a lightweight convolutional decoder. This decoder aggregates information from different hierarchical levels of the encoder to produce a high-quality image. The overall architecture is depicted in Fig.~\ref{fig: block} (c). The usage of $3\times 3$ convolutions in the decoder helps to overcome a major limitation of patch border artifacts~\cite{degae} that occur when using transformer models for image restoration. The effect is more exacerbated when using vision transformers due to its large patch size. In addition to training the above Swin transformer-based architecture, we also train two other non-generative restoration networks: NAFNet~\cite{nafnet} and PromptIR~\cite{promptir}, on the GenDS dataset. To show that the GenDS dataset benefits generative image restoration, we also include two recent diffusion-based restoration models, namely, DA-CLIP~\cite{daclip} and Diff-Plugin~\cite{diffplugin}.
\section{Experiments}
\label{sec:expts}

In this section, we provide detailed results and analysis of our method. Implementation details and dataset details can be found in the supplementary.

\begin{table*}[htbp]
\centering
\small
\caption{Quantitative comparisons of mean LPIPS and FID scores (lower is better) across within-distribution datasets for each degradation. Comparisons are shown for PromptIR~\cite{promptir}, NAFNet~\cite{nafnet}, Swin transformer, DA-CLIP~\cite{daclip} and Diff-Plugin~\cite{diffplugin} models trained with and without our GenDS dataset, along with SOTA models. Diff-Plugin\textsuperscript{\#} is the publicly available pre-trained model.}
\vskip-8pt
\label{tab: quant_within}
\begin{tabular}{lcccccc}
\toprule
\textbf{Method} & \textbf{Haze} & \textbf{Rain} & \textbf{Snow} & \textbf{Motion Blur} & \textbf{Raindrop} & \textbf{Low-light} \\
\midrule
DiffUIR          & 0.329/141.46 & 0.175/53.53 & 0.305/23.67 & 0.182/42.89 & -           & 0.551/260.26 \\
Diff-Plugin\textsuperscript{\#}       & 0.351/154.12 & 0.205/47.10 & 0.227/26.86 & 0.218/50.57 & -           & 0.464/180.67 \\
InstructIR       & 0.355/158.81 & 0.144/35.03 & -           & 0.148/31.86 & -           & 0.402/157.74 \\
AutoDIR          & 0.306/136.27 & 0.139/38.12 & -           & 0.161/33.34 & 0.195/68.09 & 0.420/155.14 \\
\midrule
PromptIR         & 0.309/141.05 & 0.097/32.61 & 0.100/18.34 & 0.163/35.79 & 0.189/84.48 & 0.421/189.87 \\
PromptIR GenDS   & 0.210/112.54 & 0.080/27.95 & 0.091/16.19 & 0.171/34.93 & 0.188/74.65 & 0.354/168.59 \\
NAFNet           & 0.190/118.22 & 0.074/21.84 & 0.067/8.20  & 0.136/28.72 & 0.085/39.91 & 0.349/172.36 \\
NAFNet GenDS     & 0.171/104.43 & 0.077/22.13 & 0.069/8.56  & 0.136/29.31 & 0.069/29.92 & 0.316/148.65 \\
Swin             & 0.244/121.92 & 0.092/24.50 & 0.080/10.80 & 0.194/40.69 & 0.097/47.09 & 0.420/187.65 \\
Swin GenDS       & 0.182/105.242 & 0.090/24.48 & 0.083/11.44 & 0.189/42.17 & 0.092/39.63 & 0.368/166.13 \\
DA-CLIP             & 0.344/147.15 & 0.114/33.32 & 0.128/43.61 & 0.198/50.03 & 0.101/40.82 & 0.538/208.47 \\
DA-CLIP GenDS       & 0.267/109.11 & 0.102/30.50 & 0.072/26.48 & 0.140/37.82 & 0.094/36.03 & 0.434/182.16 \\
Diff-Plugin             & 0.340/143.66 & 0.165/39.71 & 0.178/18.08 & 0.147/37.68 & 0.185/60.64 & 0.466/167.63 \\
Diff-Plugin GenDS       & 0.302/128.01 & 0.164/39.22 & 0.176/17.70 & 0.142/35.99 & 0.133/42.52 & 0.456/156.18 \\
\bottomrule
\end{tabular}
\end{table*}

\begin{figure*}
    \centering
    \small
    \setlength{\tabcolsep}{1pt}
    \begin{tabular}{cccccccc}
        Input&PromptIR&PromptIR GD&Swin&Swin GD&NAFNet&NAFNet GD&GT\\
         \includegraphics[height=1.5cm, width=2cm]{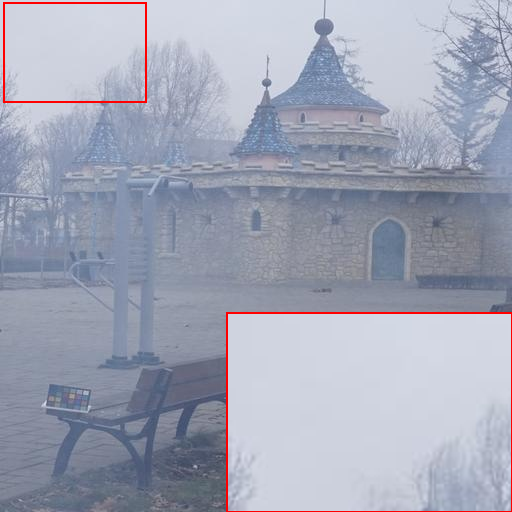}& \includegraphics[height=1.5cm, width=2cm]{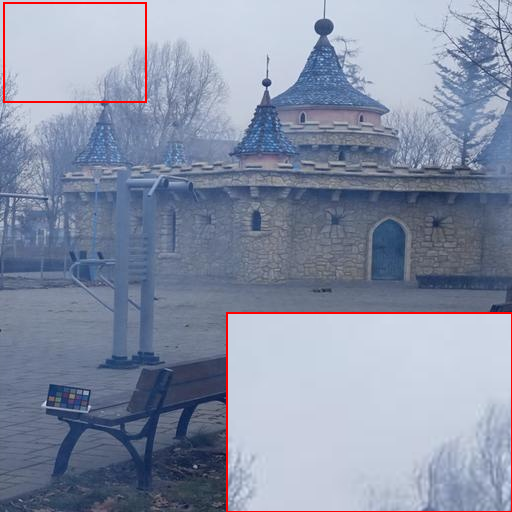}&\includegraphics[height=1.5cm, width=2cm]{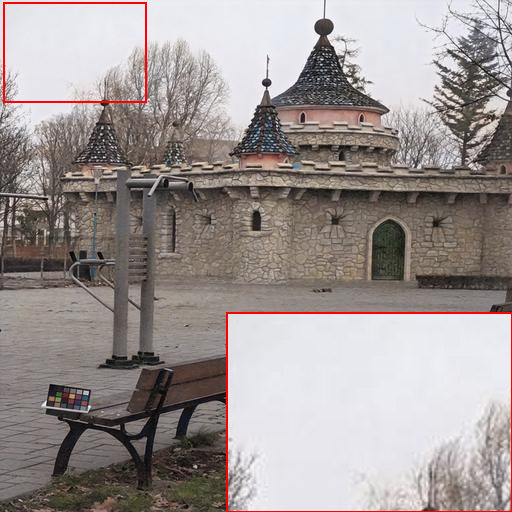}&\includegraphics[height=1.5cm, width=2cm]{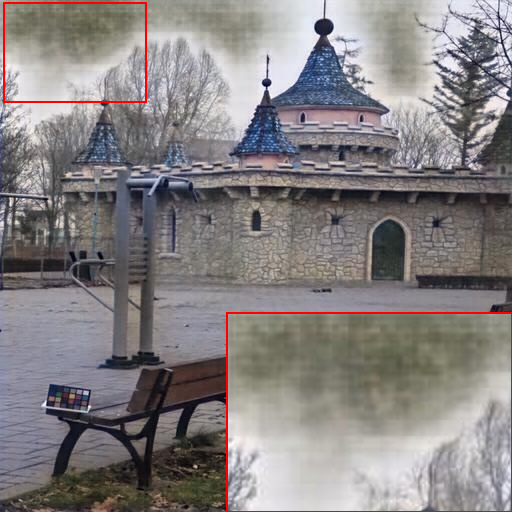}&\includegraphics[height=1.5cm, width=2cm]{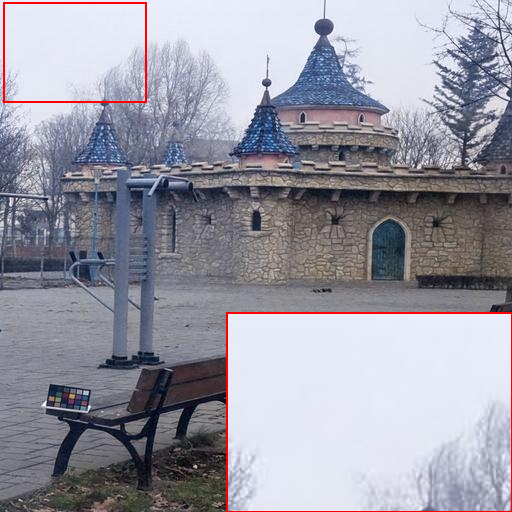}&\includegraphics[height=1.5cm, width=2cm]{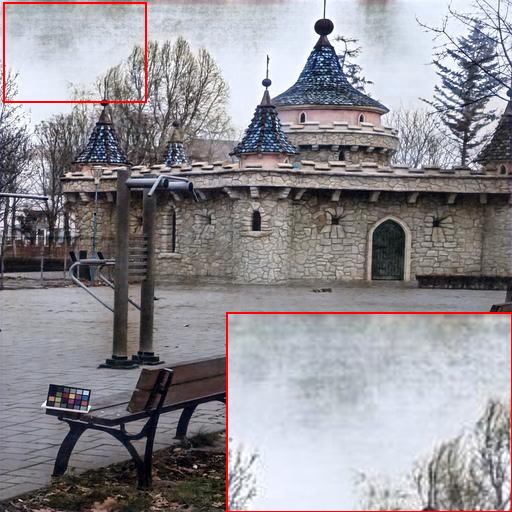}&\includegraphics[height=1.5cm, width=2cm]{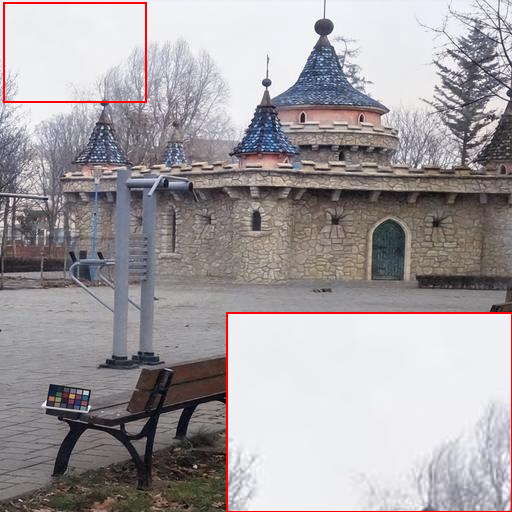}&\includegraphics[height=1.5cm, width=2cm]{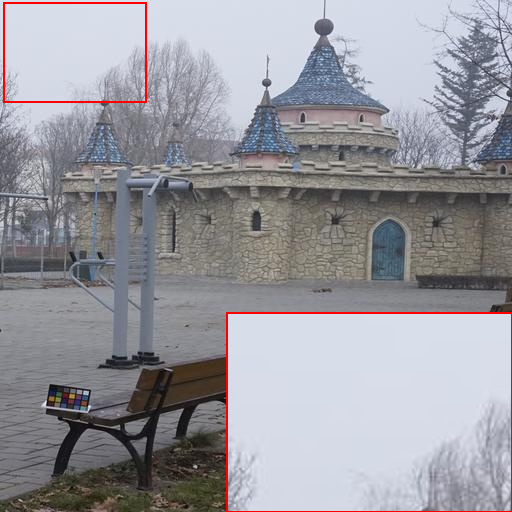} \\

         \includegraphics[height=1.5cm, width=2cm]{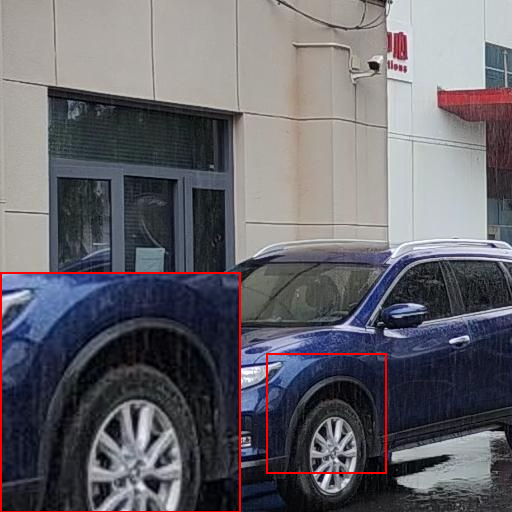}& \includegraphics[height=1.5cm, width=2cm]{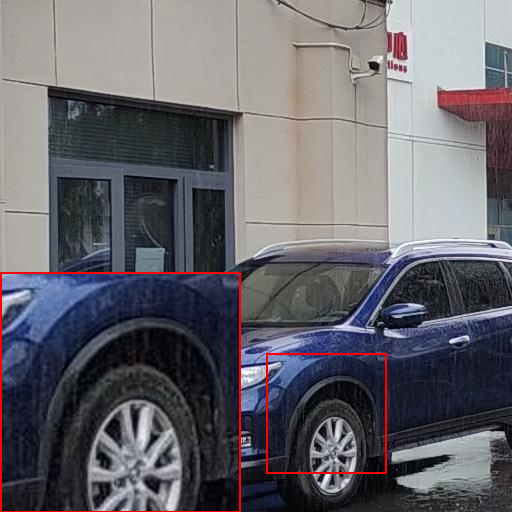}&
         \includegraphics[height=1.5cm, width=2cm]{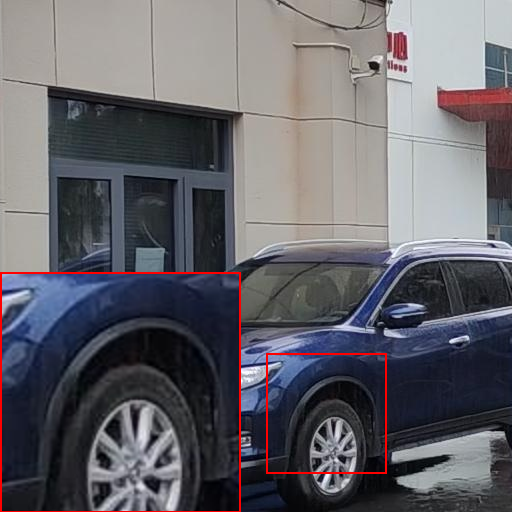}&
         \includegraphics[height=1.5cm, width=2cm]{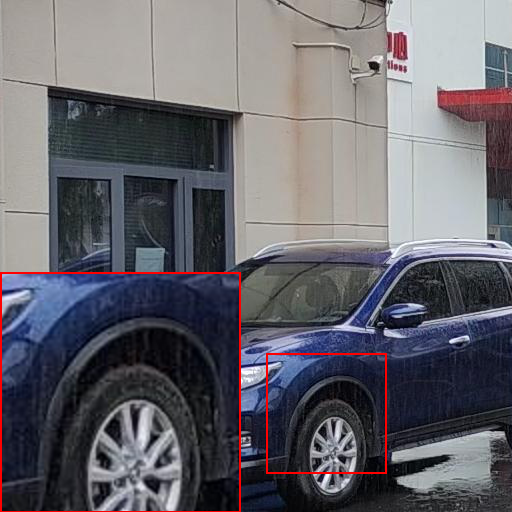}&
         \includegraphics[height=1.5cm, width=2cm]{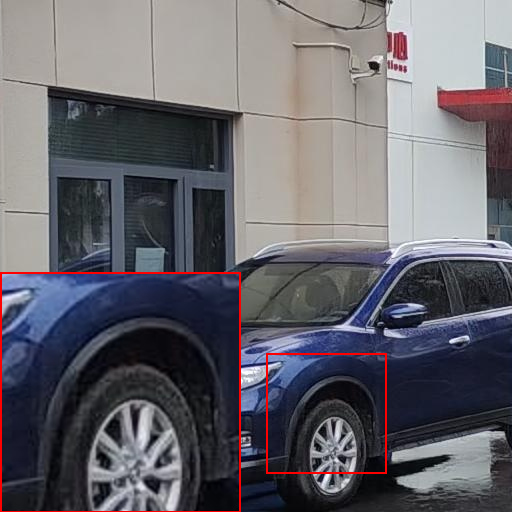}&
         \includegraphics[height=1.5cm, width=2cm]{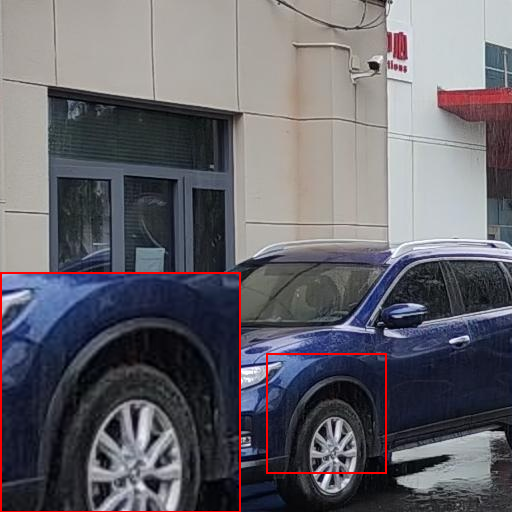}&
         \includegraphics[height=1.5cm, width=2cm]{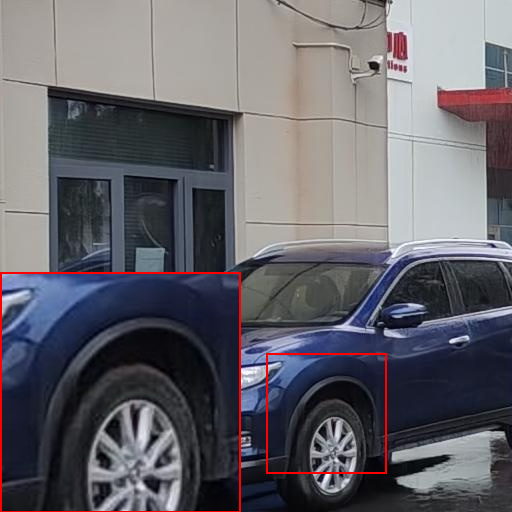}&
         \includegraphics[height=1.5cm, width=2cm]{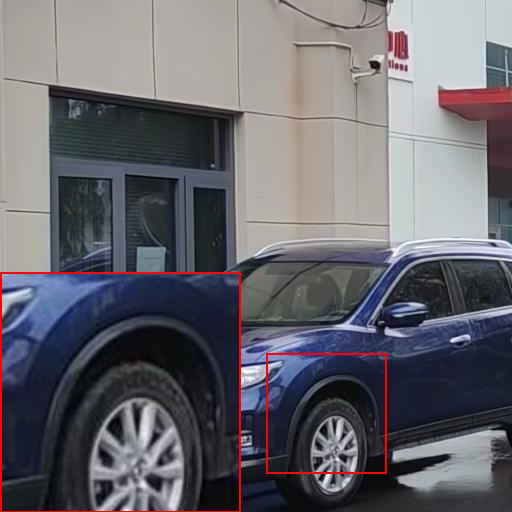}\\

         \includegraphics[height=1.5cm, width=2cm]{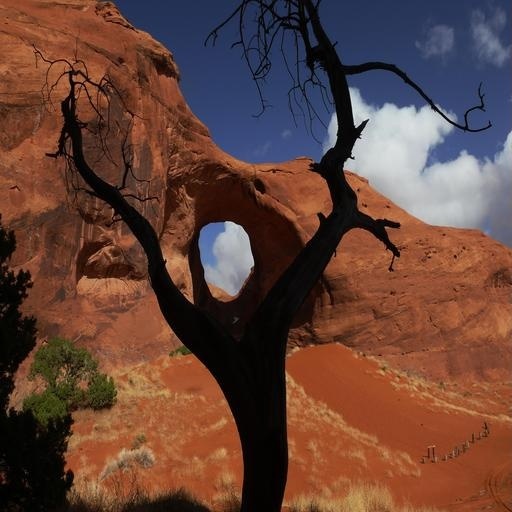}&\includegraphics[height=1.5cm, width=2cm]{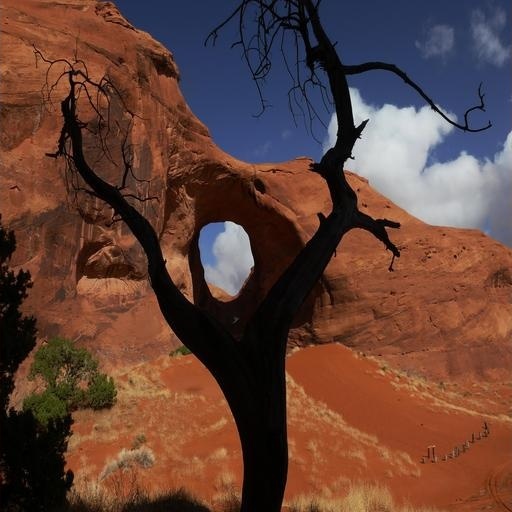}&\includegraphics[height=1.5cm, width=2cm]{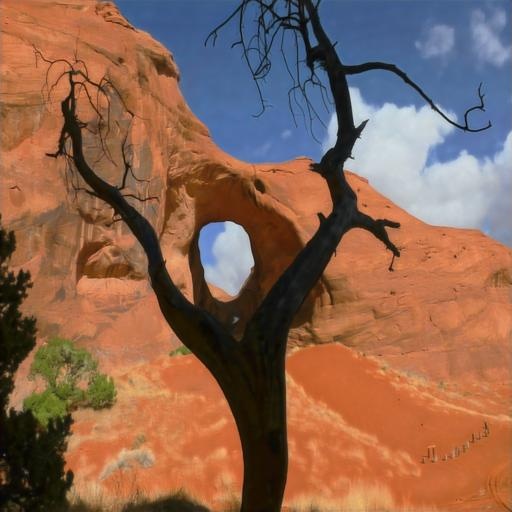}&\includegraphics[height=1.5cm, width=2cm]{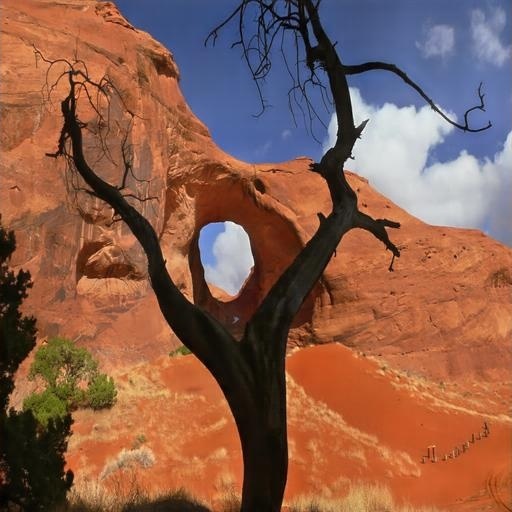}&\includegraphics[height=1.5cm, width=2cm]{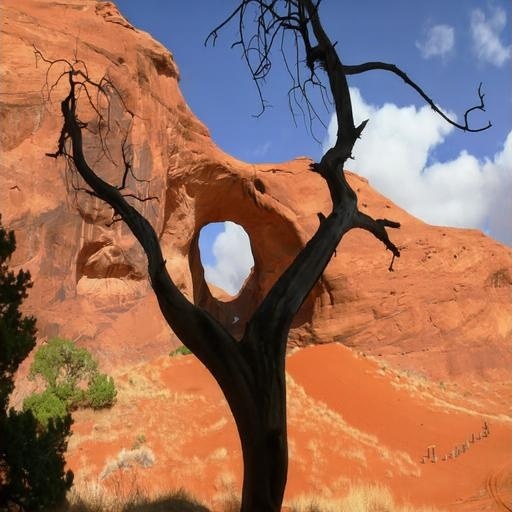}&\includegraphics[height=1.5cm, width=2cm]{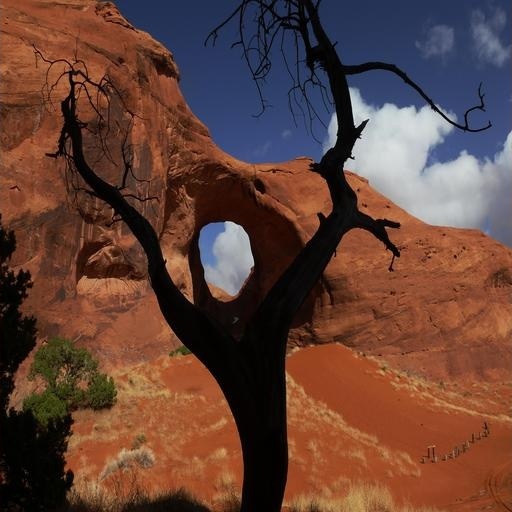}&\includegraphics[height=1.5cm, width=2cm]{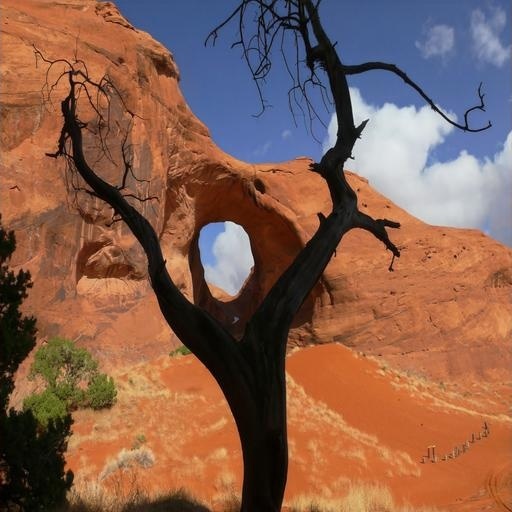}&\includegraphics[height=1.5cm, width=2cm]{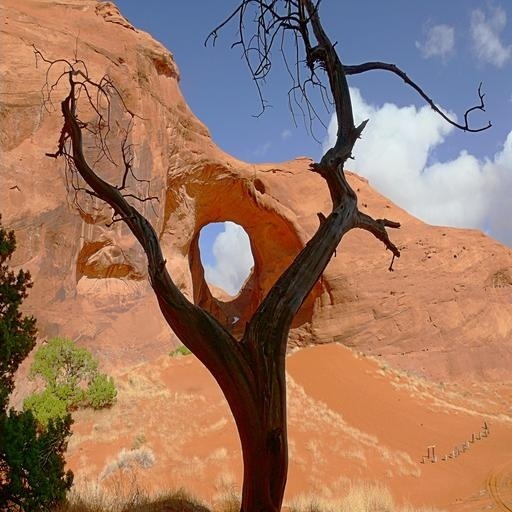}\\

         \includegraphics[height=1.5cm, width=2cm]{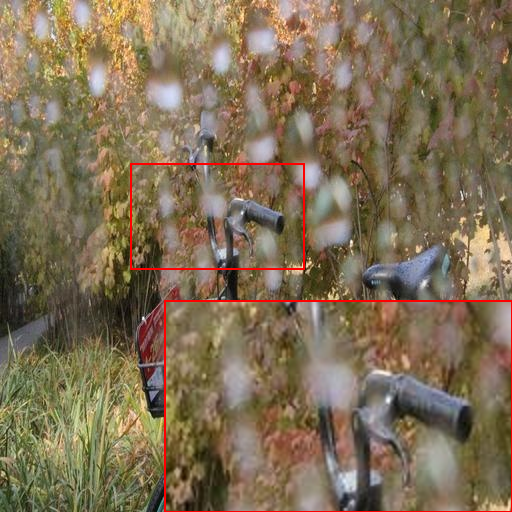}& \includegraphics[height=1.5cm, width=2cm]{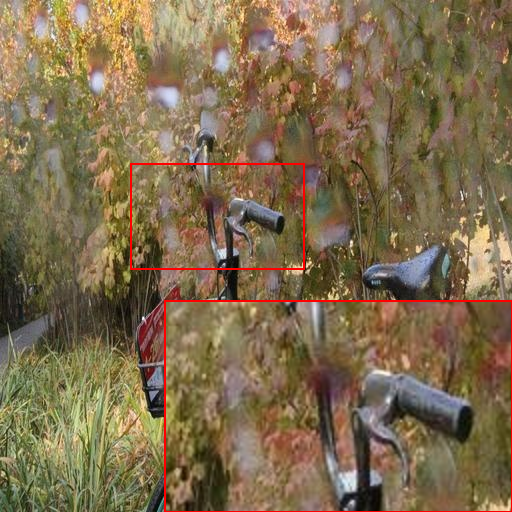}&
         \includegraphics[height=1.5cm, width=2cm]{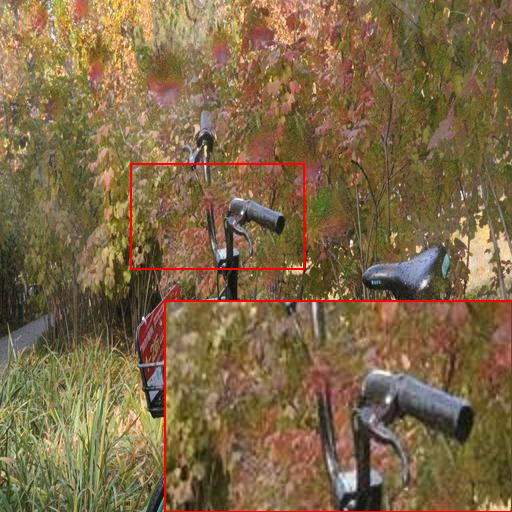}&
         \includegraphics[height=1.5cm, width=2cm]{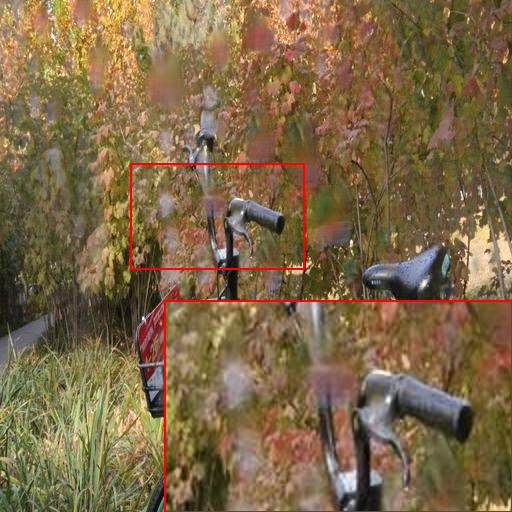}&
         \includegraphics[height=1.5cm, width=2cm]{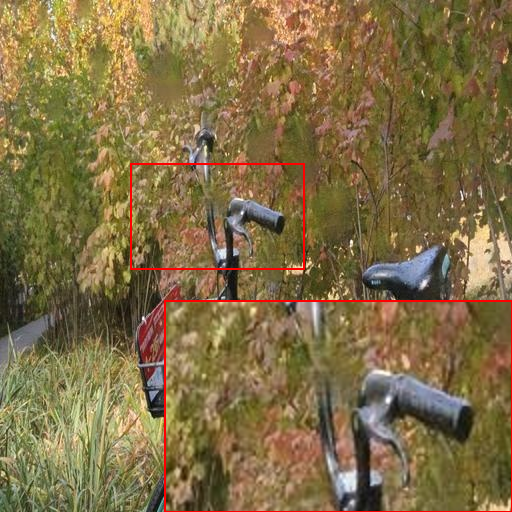}&
         \includegraphics[height=1.5cm, width=2cm]{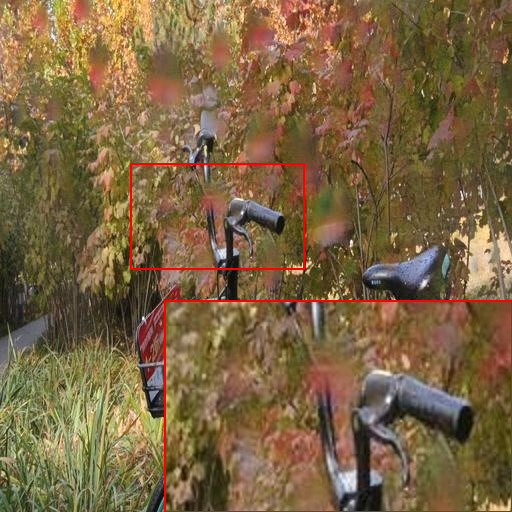}&\includegraphics[height=1.5cm, width=2cm]{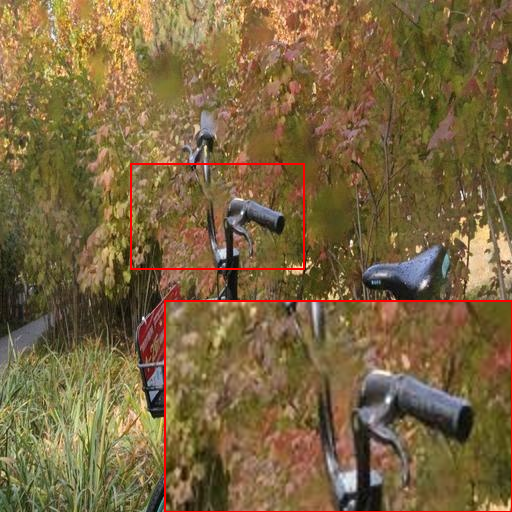}&
         \includegraphics[height=1.5cm, width=2cm]{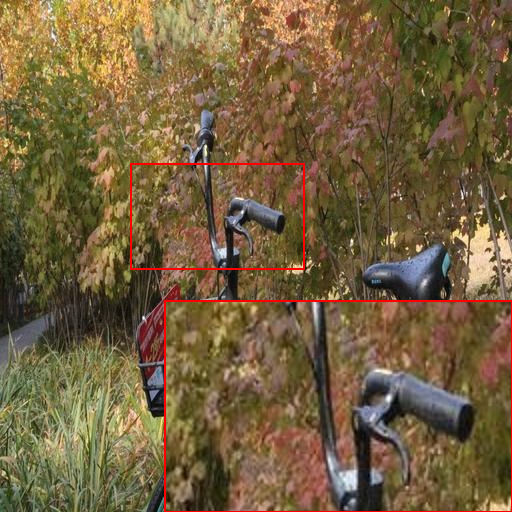}\\

         \includegraphics[height=1.5cm, width=2cm]{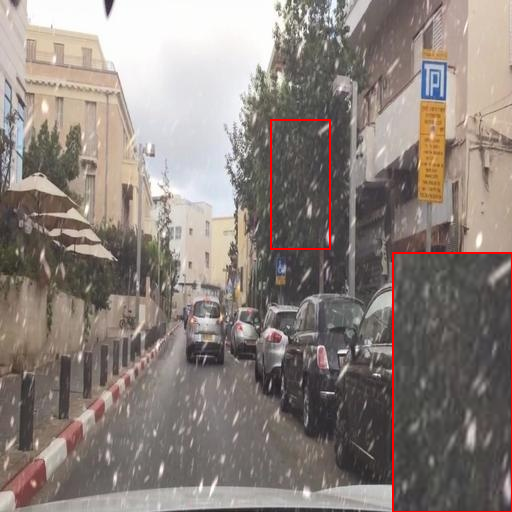}& \includegraphics[height=1.5cm, width=2cm]{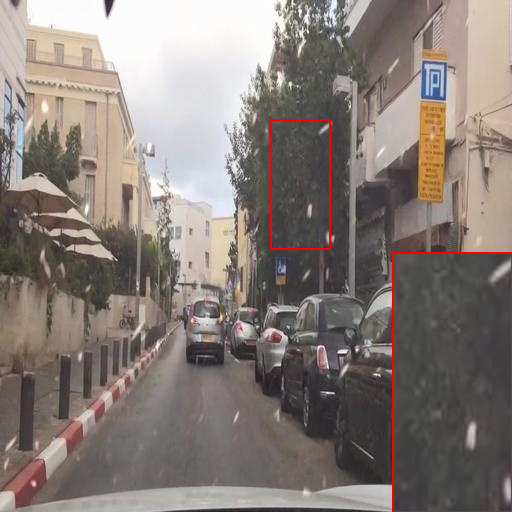}&
         \includegraphics[height=1.5cm, width=2cm]{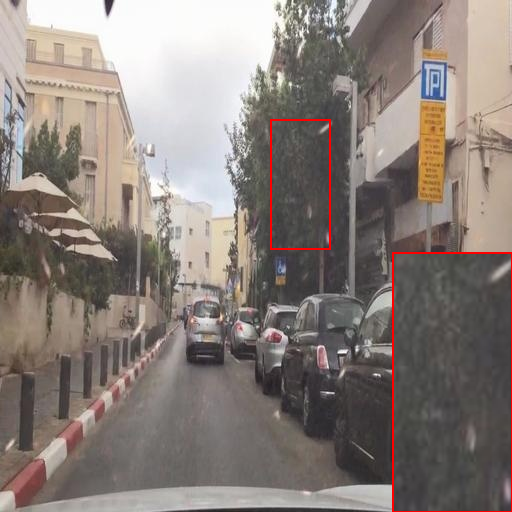}&
         \includegraphics[height=1.5cm, width=2cm]{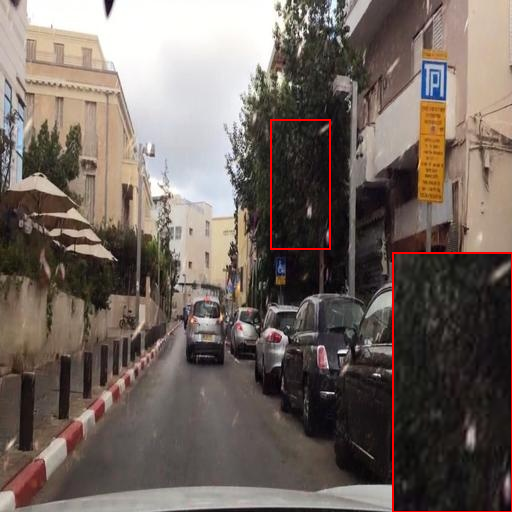}&
         \includegraphics[height=1.5cm, width=2cm]{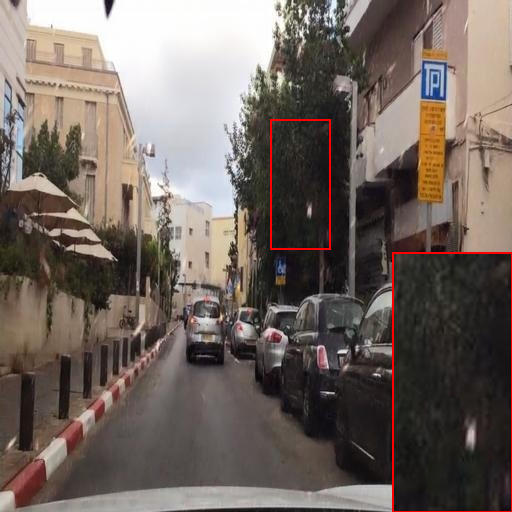}&
         \includegraphics[height=1.5cm, width=2cm]{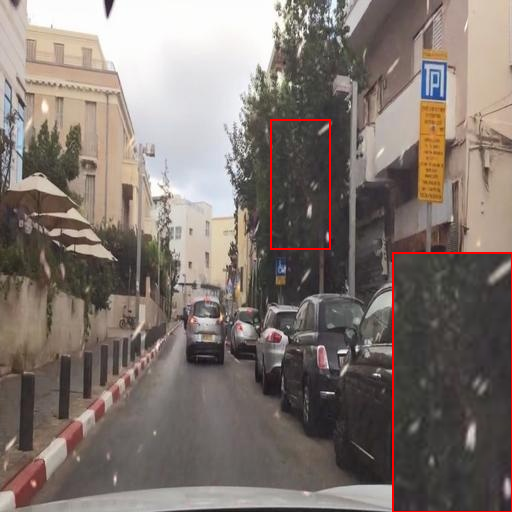}&
         \includegraphics[height=1.5cm, width=2cm]{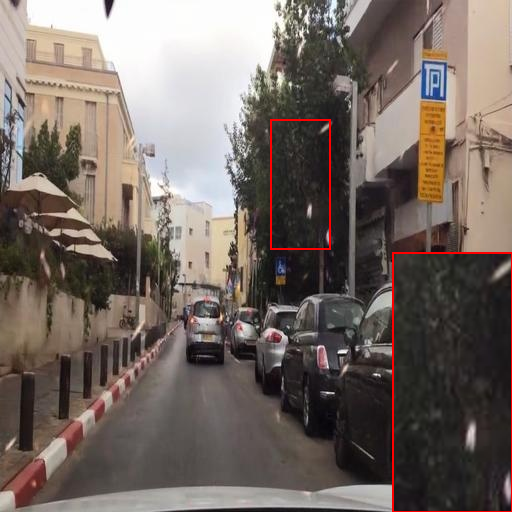}&
         \includegraphics[height=1.5cm, width=2cm]{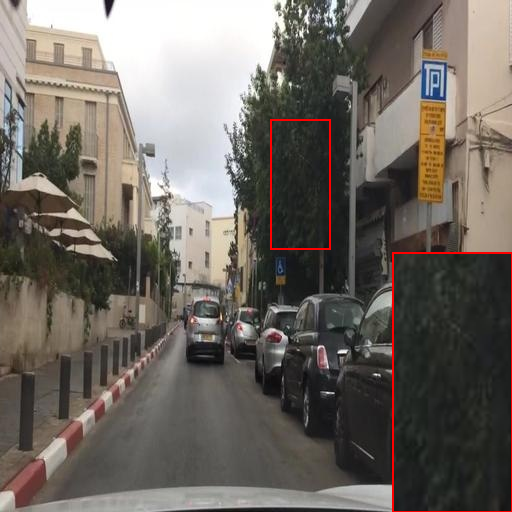}\\

         \includegraphics[height=1.5cm, width=2cm]{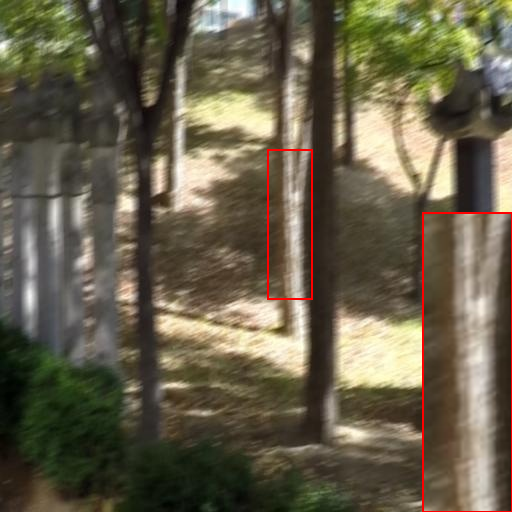}& \includegraphics[height=1.5cm, width=2cm]{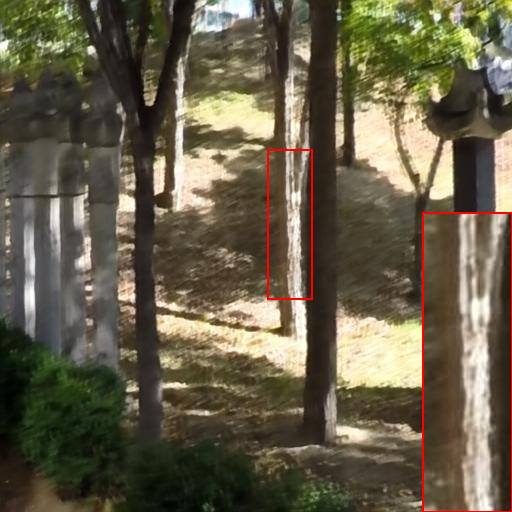}&
         \includegraphics[height=1.5cm, width=2cm]{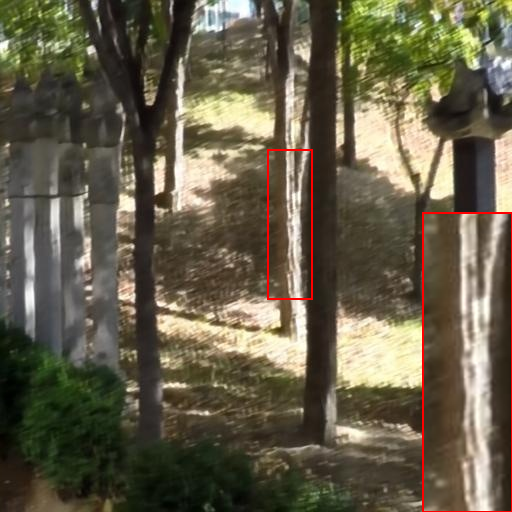}&
         \includegraphics[height=1.5cm, width=2cm]{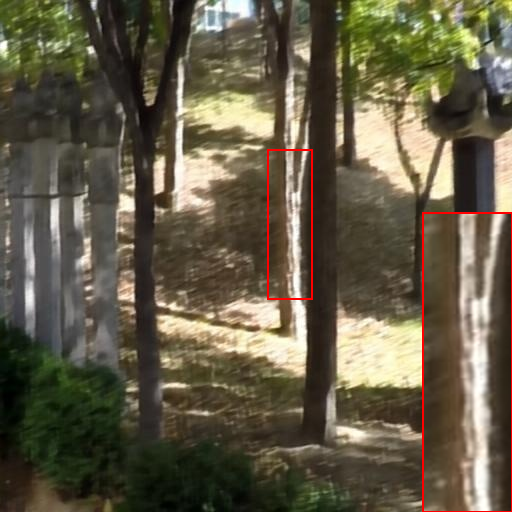}&
         \includegraphics[height=1.5cm, width=2cm]{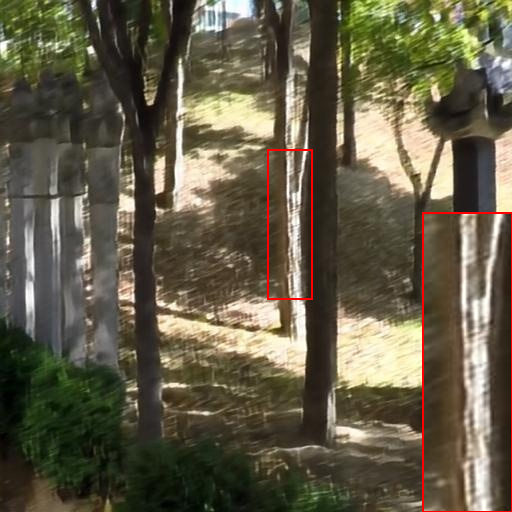}&
         \includegraphics[height=1.5cm, width=2cm]{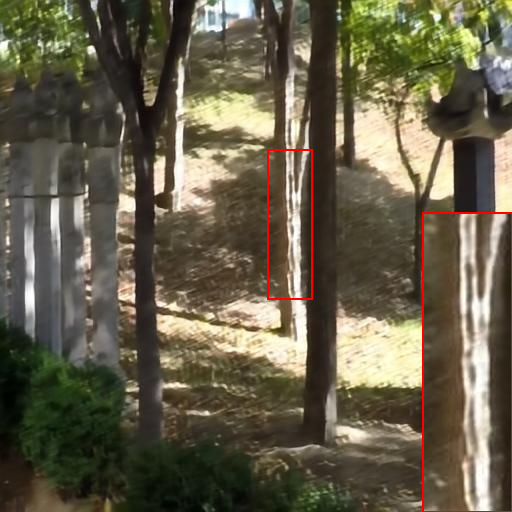}&
         \includegraphics[height=1.5cm, width=2cm]{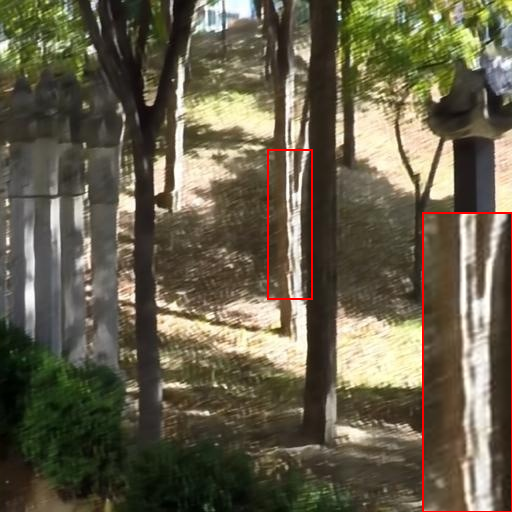}&
         \includegraphics[height=1.5cm, width=2cm]{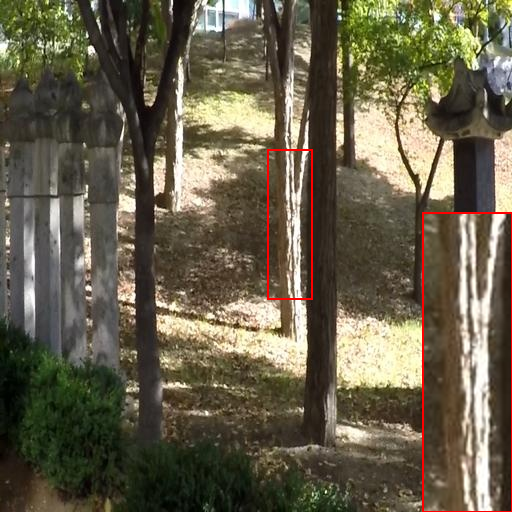}\\
         
    \end{tabular}
    \vskip-10pt
    \caption{Qualitative comparisons of image restoration models trained with and without our GenDS dataset. The suffix GD represents training with the GenDS dataset. Zoomed-in patches are provided for viewing fine details.}
    \label{fig: qual}
    \vskip-5pt
\end{figure*}

\subsection{Results}
\label{subsec: results}

To assess the impact of the GenDS dataset, we first trained NAFNet~\cite{nafnet}, PromptIR~\cite{promptir}, the Swin model (Sec.~\ref{subsec: genirmodel}), DA-CLIP~\cite{daclip} and Diff-Plugin~\cite{diffplugin} solely on existing restoration datasets, without incorporating our synthetic data. We then evaluated their performance on both within-distribution and out-of-distribution (OoD) test sets. Subsequently, we retrained the same models using the GenDS dataset and evaluated the performance. Additionally, we compared their performance against state-of-the-art (SOTA) AIOR models, namely, DiffUIR~\cite{diffuir}, Diff-Plugin~\cite{diffplugin} (pre-trained model), InstructIR~\cite{instructir} and AutoDIR~\cite{autodir}.

\textbf{Quantitative comparisons. }Due to space constraints, we present quantitative comparisons using only the LPIPS and FID metrics (following~\cite{diffplugin}). Table~\ref{tab: quant_ood} presents these scores for OoD test sets across all six degradations. We observe that PromptIR, NAFNet, the Swin model, DA-CLIP and Diff-Plugin exhibit significant improvements in OoD performance when trained on the GenDS dataset. Motion blur performance remains nearly the same even after training with the GenDS dataset. Since motion blur already contains sufficient real data with diverse scenes (see Fig.~\ref{fig: datastats}), introducing more synthetic data does not impact performance on real OoD samples. This highlights the importance of diverse high-quality data for generalizable AIOR. 

Furthermore, the results indicate that the synthetic data generated by GenDeg aids in bridging the domain gap with OoD samples. In certain instances, SOTA methods (upper portion of Table~\ref{tab: quant_ood}) outperform our models on specific OoD datasets (e.g., AutoDIR raindrop removal on RainDS). However, it is important to note that our models serve as simple baselines compared to the more complex SOTA architectures, and that SOTA methods do not consistently perform well for OoD datasets across degradations. Furthermore, SOTA architectures could further enhance their OoD performance by training with our GenDS dataset, as evidenced by the improvements observed for Diff-Plugin.

Table~\ref{tab: quant_within} presents the mean within-distribution performance for each degradation. Interestingly, the performance remains almost identical even after training with our GenDS dataset. Moreover, the within-distribution performance shows substantial improvements for haze, low-light and raindrop degradations. This improvement is likely due to the GenDS dataset effectively addressing the limited scene diversity present in existing datasets for these degradations. Detailed quantitative results (including PSNR and SSIM metrics) are available in the supplementary material.

\textbf{Qualitative comparisons. }Fig.~\ref{fig: qual} shows qualitative results from one OoD test set per degradation for three top-performing models (PromptIR, NAFNet and the Swin model). Comparisons with SOTA are in the supplementary. Models trained with GenDS dataset consistently yield the best results. Notably, the enhanced images often contain richer colors than the ground truth (see first row), which can lower PSNR and SSIM scores. Thus, LPIPS and FID scores are more reliable for testing the OoD performance. 
%We provide qualitative results from one OoD test set for each degradation in Fig.~\ref{fig: qual}. The comparisons are provided for three top-performing approaches (PromptIR, NAFNet and the Swin model). 
These results demonstrate that the synthetic data generated by GenDeg enhances the generalization of AIOR models.

\subsection{Analysis}
\label{subsec: analysis}
In this section, we use our generated synthetic data to conduct various insightful analyses.

\begin{figure}
    \centering
    \includegraphics[width=1\linewidth]% {figs/analysis/dual_y_axes_plot.png}
    {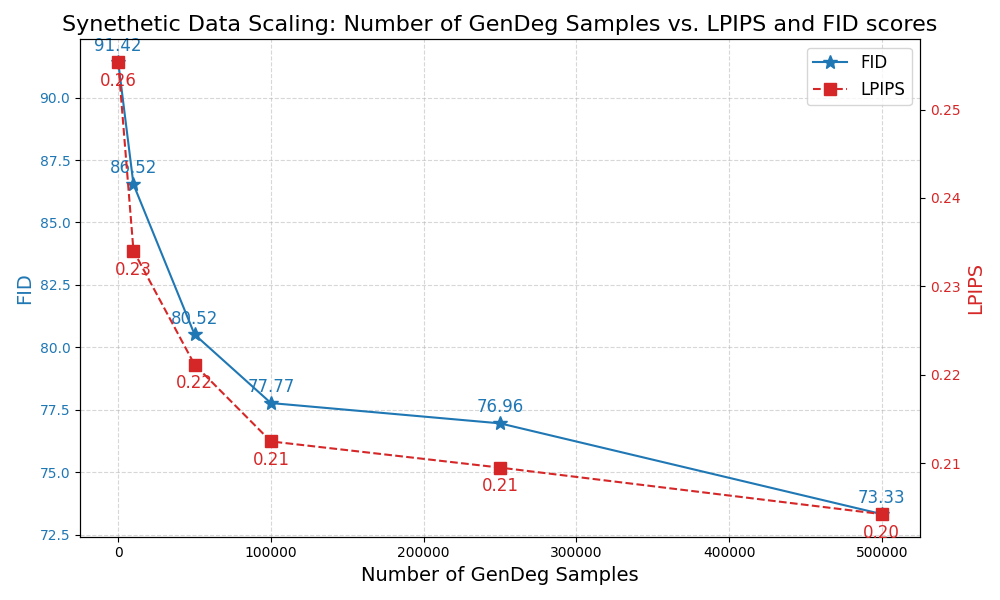}
    \vskip-5pt
    \caption{Effect of scaling number of GenDeg samples augmented with real data on OoD performance (LPIPS and FID).}
    \label{fig: scaling}
    \vskip-5pt
    
\end{figure}

\textbf{Synthetic Data Scaling. }We analyze the impact of progressively adding synthetic data generated by GenDeg to existing real data on out-of-distribution (OoD) performance, utilizing the Swin transformer model. Fig.~\ref{fig: scaling} illustrates the variation in mean OoD LPIPS and FID scores with increasing synthetic data. There is substantial OoD performance improvements with the addition of up to $100$k synthetic samples, after which the performance improvement is reduced. Due to limited computational resources, we were unable to scale the synthetic dataset beyond $500$k samples.

\textbf{Generated degradation diversity. }GenDeg-generated data significantly aids in improving OoD performance of restoration models. This improvement is primarily due to the enhanced scene diversity (as shown in Fig.~\ref{fig: datastats}) in our dataset and the variety of degradation patterns produced by GenDeg. To illustrate the diversity in degradation patterns, we utilize degradation-aware CLIP (DA-CLIP~\cite{daclip}), a robust CLIP model trained to extract degradation-specific features from images. Fig.~\ref{fig: tsne} presents a t-SNE visualization of the DA-CLIP degradation embeddings obtained from hazy samples in existing training datasets, our GenDS dataset, and the OoD test sets. The visualization reveals a clear gap between features of the training dataset and OoD test sets. This indicates that the degradation patterns in these datasets are different, hindering generalization. GenDS dataset bridges this gap by introducing numerous samples that resemble those in the OoD test sets, thereby enhancing generalization. Note that GenDeg was \textit{never} trained on the OoD test sets. Furthermore, the t-SNE plot showcases the diversity of degradation patterns produced by our model, as evidenced by our samples spanning a wide area.

\textbf{Domain gap with real data. }We examine the domain gap between existing datasets and GenDeg-generated data by training the Swin model exclusively on GenDeg synthesized data. Table~\ref{tab: synthetic} provides the mean LPIPS and FID scores for both within-distribution and OoD testing across all degradations. Training solely with GenDeg data lowers within-distribution performance compared to training on existing datasets due to the domain gap caused by factors such as alignment discrepancies between diffusion-generated samples and corresponding clean samples. However, utilizing both existing and GenDeg synthesized data, i.e., GenDS dataset, enhances performance as the restoration model benefits from diverse degradation patterns and scenes while maintaining performance on existing data. 

The model trained only on existing datasets performs worse on OoD data than within-distribution. Training solely on GenDeg data improves OoD performance demonstrating that the diversity in scenes and degradation patterns enhances generalization. Nevertheless, the best performance is achieved when training on the GenDS dataset.

%For this experiment, we provide mean of the within distribution and OoD performance across all degradations. 

\begin{table}[t]
    \centering
    \small
    \setlength{\tabcolsep}{1pt}
    \caption{LPIPS/FID scores for analyzing the performance difference between training on solely existing data, solely GenDeg data, and both existing and GenDeg data (GenDS dataset).}
    \vskip-8pt
    \begin{tabular}{c|c|c|c}
         \textbf{Setting}&\textbf{Existing data}&\textbf{GenDeg data}&\textbf{GenDS data}  \\ \hline
         Within-distribution&0.1879/72.10&0.2358/81.08&0.1669/64.84\\
         OoD&0.211/59.36&0.207/61.71&0.1694/47.99
    \end{tabular}

    \label{tab: synthetic}
    % \vskip-8pt
\end{table}

\begin{figure}
    \centering
    \small
    \setlength{\tabcolsep}{1pt}
    \begin{tabular}{cccc}
        Image, $\sigma: 0.12$&$\mu:0.1$&$\mu:0.2$&$\mu:0.27$\\
         \includegraphics[height=2cm, width=2cm]{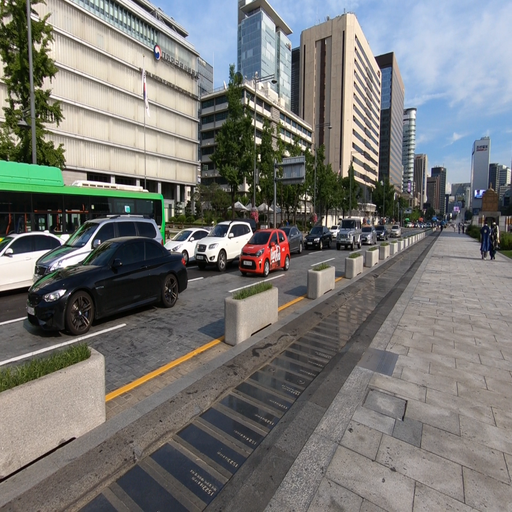}&\includegraphics[height=2cm, width=2cm]{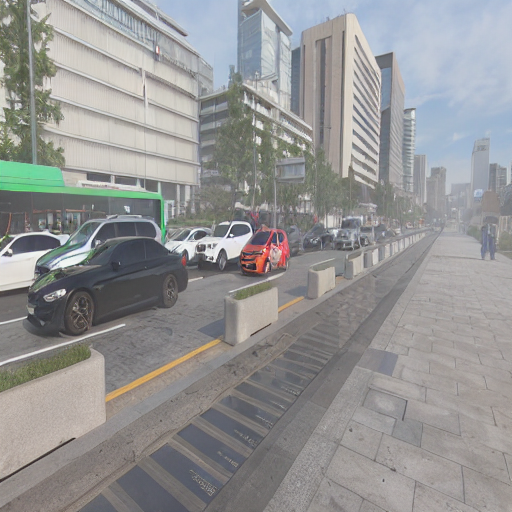}&\includegraphics[height=2cm, width=2cm]{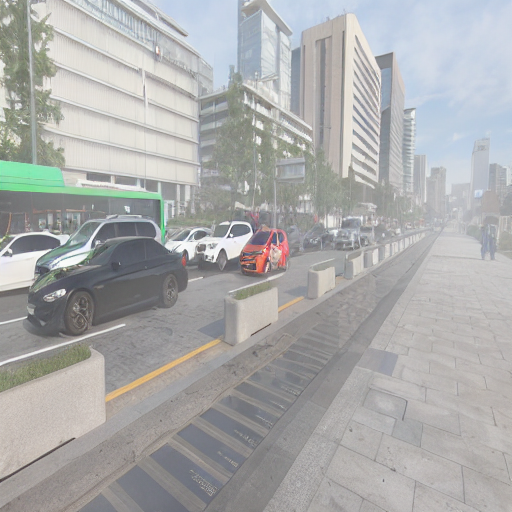}&\includegraphics[height=2cm, width=2cm]{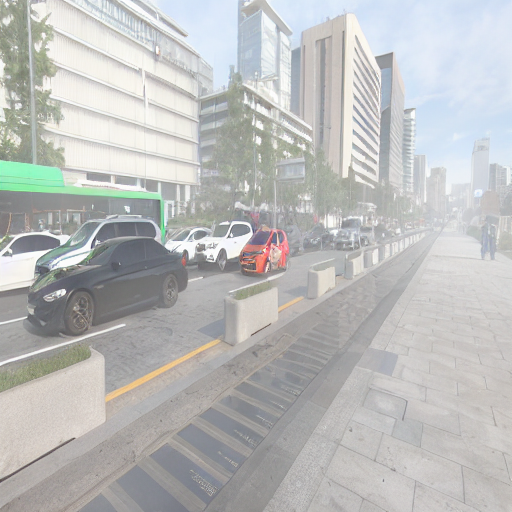}  \\

         Image, $\mu:0.15$&$\sigma:0.15$&$\sigma:0.17$&$\sigma:0.2$\\
         \includegraphics[height=2cm, width=2cm]{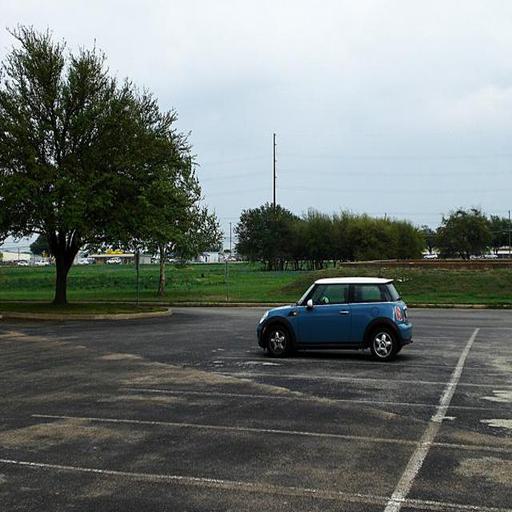}&\includegraphics[height=2cm, width=2cm]{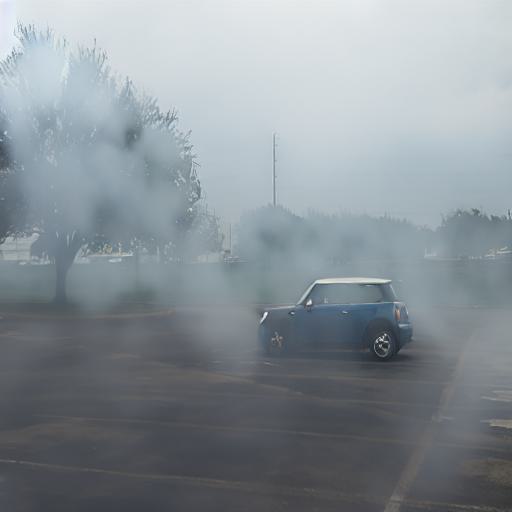}&\includegraphics[height=2cm, width=2cm]{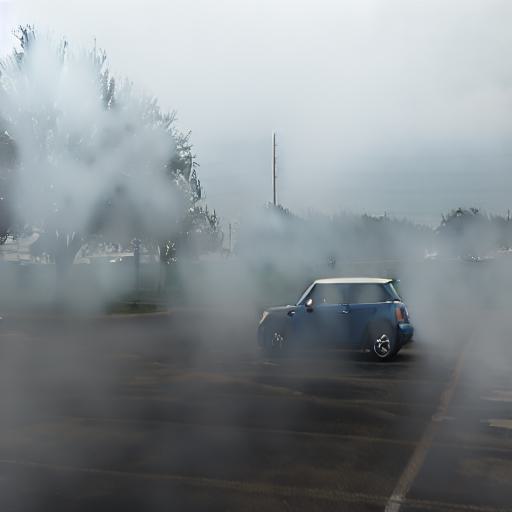}&\includegraphics[height=2cm, width=2cm]{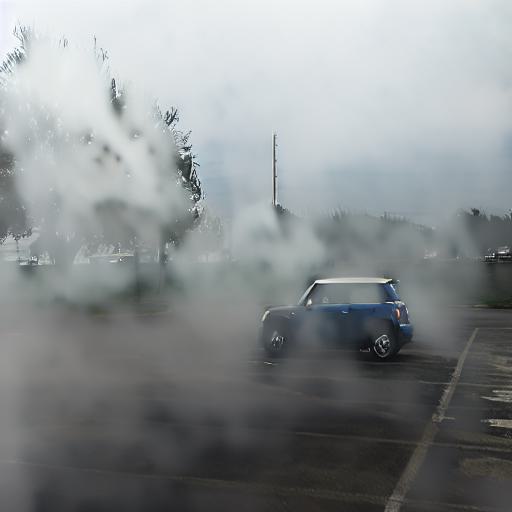}
    \end{tabular}
    \vskip-5pt
    \caption{Effect of varying $\mu$ and $\sigma$ in our GenDeg framework for the degradation of haze.}
    \label{fig: musigma}
    \vskip-8pt
\end{figure}

\textbf{Effect of $\mu$ and $\sigma$. }GenDeg enables control over the intensity and variations in the generated degradations, thereby enhancing diversity. In Fig.~\ref{fig: musigma}, we illustrate the effect of varying $\mu$ and $\sigma$ on two images for the synthesized degradation of haze. When $\mu$ is increased, the intensity of haze in the image expectedly increases. $\sigma$ controls the variation of haze in the image. For $\sigma=0.15$, the non-homogenous haze (similar to NH-Haze~\cite{nhhaze}) is spread throughout the image. As $\sigma$ increases, the spread of haze becomes more localized with higher intensity as seen from the figure. 
% Fig.~\ref{fig: musigma} illustrates the same by first fixing the value of $\sigma$ and varying $\mu$ and then fixing the value of $\mu$ and varying $\sigma$. 

\section{Conclusions}
\label{sec:conclusions}
In this paper, we addressed the important problem of generalization in All-In-One Restoration (AIOR) models. Toward this aim, we introduced GenDeg, a novel diffusion model–based framework for synthesizing diverse degradations on clean images, offering fine-grained control over degradation characteristics. Utilizing GenDeg, we generated over $550$k degraded samples encompassing a wide range of scenes and degradations. Training AIOR models with both existing and GenDeg data yielded significant improvements in out-of-distribution performance. Our work suggests a promising research direction for addressing generalization challenges in AIOR, aiding in the development of more robust restoration models.

\section*{Acknowledgments}

This work is supported by the Intelligence Advanced Research Projects Activity (IARPA) via Department of Interior/ Interior Business Center (DOI/IBC) contract number 140D0423C0076. The U.S. Government is authorized to reproduce and distribute reprints for Governmental purposes notwithstanding any copyright annotation thereon. Disclaimer: The views and conclusions contained herein are those of the authors and should not be interpreted as necessarily representing the official policies or endorsements, either expressed or implied, of IARPA, DOI/IBC, or the U.S. Government.

{
   \small
   \bibliographystyle{ieeenat_fullname}
   \bibliography{main}

\begin{thebibliography}{62}
\providecommand{\natexlab}[1]{#1}
\providecommand{\url}[1]{\texttt{#1}}
\expandafter\ifx\csname urlstyle\endcsname\relax
  \providecommand{\doi}[1]{doi: #1}\else
  \providecommand{\doi}{doi: \begingroup \urlstyle{rm}\Url}\fi

\bibitem[Ancuti et~al.(2018)Ancuti, Ancuti, Timofte, and De~Vleeschouwer]{ohaze}
Codruta~O Ancuti, Cosmin Ancuti, Radu Timofte, and Christophe De~Vleeschouwer.
\newblock O-haze: a dehazing benchmark with real hazy and haze-free outdoor images.
\newblock In \emph{Proceedings of the IEEE conference on computer vision and pattern recognition workshops}, pages 754--762, 2018.

\bibitem[Ancuti et~al.(2020)Ancuti, Ancuti, and Timofte]{nhhaze}
Codruta~O Ancuti, Cosmin Ancuti, and Radu Timofte.
\newblock Nh-haze: An image dehazing benchmark with non-homogeneous hazy and haze-free images.
\newblock In \emph{Proceedings of the IEEE/CVF conference on computer vision and pattern recognition workshops}, pages 444--445, 2020.

\bibitem[Azizi et~al.(2023)Azizi, Kornblith, Saharia, Norouzi, and Fleet]{diffclass2}
Shekoofeh Azizi, Simon Kornblith, Chitwan Saharia, Mohammad Norouzi, and David~J Fleet.
\newblock Synthetic data from diffusion models improves imagenet classification.
\newblock \emph{arXiv preprint arXiv:2304.08466}, 2023.

\bibitem[Bansal and Grover(2023)]{diffclass1}
Hritik Bansal and Aditya Grover.
\newblock Leaving reality to imagination: Robust classification via generated datasets.
\newblock \emph{arXiv preprint arXiv:2302.02503}, 2023.

\bibitem[Brooks et~al.(2023)Brooks, Holynski, and Efros]{ip2p}
Tim Brooks, Aleksander Holynski, and Alexei~A Efros.
\newblock Instructpix2pix: Learning to follow image editing instructions.
\newblock In \emph{Proceedings of the IEEE/CVF Conference on Computer Vision and Pattern Recognition}, pages 18392--18402, 2023.

\bibitem[Cai et~al.(2018)Cai, Gu, and Zhang]{sice}
Jianrui Cai, Shuhang Gu, and Lei Zhang.
\newblock Learning a deep single image contrast enhancer from multi-exposure images.
\newblock \emph{IEEE Transactions on Image Processing}, 27\penalty0 (4):\penalty0 2049--2062, 2018.

\bibitem[Cantor(1978)]{hazemodel1}
A. Cantor.
\newblock Optics of the atmosphere--scattering by molecules and particles.
\newblock \emph{IEEE Journal of Quantum Electronics}, 14\penalty0 (9):\penalty0 698--699, 1978.

\bibitem[Chen et~al.(2023)Chen, Ren, Gu, Wu, Lu, Cai, and Zhu]{rsvd}
Haoyu Chen, Jingjing Ren, Jinjin Gu, Hongtao Wu, Xuequan Lu, Haoming Cai, and Lei Zhu.
\newblock Snow removal in video: A new dataset and a novel method.
\newblock In \emph{2023 IEEE/CVF International Conference on Computer Vision (ICCV)}, pages 13165--13176. IEEE, 2023.

\bibitem[Chen et~al.(2024)Chen, Huang, Lv, Cui, Chen, and Wei]{textdiffuser}
Jingye Chen, Yupan Huang, Tengchao Lv, Lei Cui, Qifeng Chen, and Furu Wei.
\newblock Textdiffuser: Diffusion models as text painters.
\newblock \emph{Advances in Neural Information Processing Systems}, 36, 2024.

\bibitem[Chen et~al.(2022{\natexlab{a}})Chen, Chu, Zhang, and Sun]{nafnet}
Liangyu Chen, Xiaojie Chu, Xiangyu Zhang, and Jian Sun.
\newblock Simple baselines for image restoration.
\newblock In \emph{European conference on computer vision}, pages 17--33. Springer, 2022{\natexlab{a}}.

\bibitem[Chen et~al.(2020)Chen, Fang, Ding, Tsai, and Kuo]{snow2}
Wei-Ting Chen, Hao-Yu Fang, Jian-Jiun Ding, Cheng-Che Tsai, and Sy-Yen Kuo.
\newblock Jstasr: Joint size and transparency-aware snow removal algorithm based on modified partial convolution and veiling effect removal.
\newblock Berlin, Heidelberg, 2020. Springer-Verlag.

\bibitem[Chen et~al.(2022{\natexlab{b}})Chen, Huang, Tsai, Yang, Ding, and Kuo]{tkmc}
Wei-Ting Chen, Zhi-Kai Huang, Cheng-Che Tsai, Hao-Hsiang Yang, Jian-Jiun Ding, and Sy-Yen Kuo.
\newblock Learning multiple adverse weather removal via two-stage knowledge learning and multi-contrastive regularization: Toward a unified model.
\newblock 2022{\natexlab{b}}.

\bibitem[Conde et~al.(2025)Conde, Geigle, and Timofte]{instructir}
Marcos~V Conde, Gregor Geigle, and Radu Timofte.
\newblock Instructir: High-quality image restoration following human instructions.
\newblock In \emph{European Conference on Computer Vision}, pages 1--21. Springer, 2025.

\bibitem[Deng et~al.(2009)Deng, Dong, Socher, Li, Li, and Fei-Fei]{imagenet}
Jia Deng, Wei Dong, Richard Socher, Li-Jia Li, Kai Li, and Li Fei-Fei.
\newblock Imagenet: A large-scale hierarchical image database.
\newblock In \emph{2009 IEEE conference on computer vision and pattern recognition}, pages 248--255. Ieee, 2009.

\bibitem[Guo et~al.(2023)Guo, Xiao, Chang, Deng, and Yan]{lhprain}
Yun Guo, Xueyao Xiao, Yi Chang, Shumin Deng, and Luxin Yan.
\newblock From sky to the ground: A large-scale benchmark and simple baseline towards real rain removal.
\newblock In \emph{Proceedings of the IEEE/CVF International Conference on Computer Vision (ICCV)}, pages 12097--12107, 2023.

\bibitem[He et~al.(2009)He, Sun, and Tang]{early1}
Kaiming He, Jian Sun, and Xiaoou Tang.
\newblock Single image haze removal using dark channel prior.
\newblock In \emph{2009 IEEE Conference on Computer Vision and Pattern Recognition}, pages 1956--1963, 2009.

\bibitem[Jiang et~al.(2024)Jiang, Zuo, Wu, Jiang, and Liu]{surveygenir}
Junjun Jiang, Zengyuan Zuo, Gang Wu, Kui Jiang, and Xianming Liu.
\newblock A survey on all-in-one image restoration: Taxonomy, evaluation and future trends.
\newblock \emph{arXiv preprint arXiv:2410.15067}, 2024.

\bibitem[Jiang et~al.(2023)Jiang, Zhang, Xue, and Gu]{autodir}
Yitong Jiang, Zhaoyang Zhang, Tianfan Xue, and Jinwei Gu.
\newblock Autodir: Automatic all-in-one image restoration with latent diffusion.
\newblock \emph{arXiv preprint arXiv:2310.10123}, 2023.

\bibitem[Kang et~al.(2012)Kang, Lin, and Fu]{early3}
Li-Wei Kang, Chia-Wen Lin, and Yu-Hsiang Fu.
\newblock Automatic single-image-based rain streaks removal via image decomposition.
\newblock \emph{IEEE Transactions on Image Processing}, 21\penalty0 (4):\penalty0 1742--1755, 2012.

\bibitem[Kingma(2013)]{vae}
Diederik~P Kingma.
\newblock Auto-encoding variational bayes.
\newblock \emph{arXiv preprint arXiv:1312.6114}, 2013.

\bibitem[Kirillov et~al.(2023)Kirillov, Mintun, Ravi, Mao, Rolland, Gustafson, Xiao, Whitehead, Berg, Lo, et~al.]{sam}
Alexander Kirillov, Eric Mintun, Nikhila Ravi, Hanzi Mao, Chloe Rolland, Laura Gustafson, Tete Xiao, Spencer Whitehead, Alexander~C Berg, Wan-Yen Lo, et~al.
\newblock Segment anything.
\newblock In \emph{Proceedings of the IEEE/CVF International Conference on Computer Vision}, pages 4015--4026, 2023.

\bibitem[Kong et~al.(2024)Kong, Gu, Liu, Zhang, Chen, Qiao, and Dong]{preliminarygenir}
Xiangtao Kong, Jinjin Gu, Yihao Liu, Wenlong Zhang, Xiangyu Chen, Yu Qiao, and Chao Dong.
\newblock A preliminary exploration towards general image restoration.
\newblock \emph{arXiv preprint arXiv:2408.15143}, 2024.

\bibitem[Li et~al.(2019)Li, Ren, Fu, Tao, Feng, Zeng, and Wang]{reside}
Boyi Li, Wenqi Ren, Dengpan Fu, Dacheng Tao, Dan Feng, Wenjun Zeng, and Zhangyang Wang.
\newblock Benchmarking single-image dehazing and beyond.
\newblock \emph{IEEE Transactions on Image Processing}, 28\penalty0 (1):\penalty0 492--505, 2019.

\bibitem[Li et~al.(2023)Li, Li, Savarese, and Hoi]{blip2}
Junnan Li, Dongxu Li, Silvio Savarese, and Steven Hoi.
\newblock Blip-2: Bootstrapping language-image pre-training with frozen image encoders and large language models.
\newblock In \emph{International conference on machine learning}, pages 19730--19742. PMLR, 2023.

\bibitem[Li et~al.(2020)Li, Tan, and Cheong]{nas}
Ruoteng Li, Robby~T. Tan, and Loong-Fah Cheong.
\newblock All in one bad weather removal using architectural search.
\newblock In \emph{2020 IEEE/CVF Conference on Computer Vision and Pattern Recognition (CVPR)}, pages 3172--3182, 2020.

\bibitem[Li et~al.(2022)Li, Zhang, Zhang, Huang, Tian, and Tao]{realrain1k}
Wei Li, Qiming Zhang, Jing Zhang, Zhen Huang, Xinmei Tian, and Dacheng Tao.
\newblock Toward real-world single image deraining: A new benchmark and beyond.
\newblock \emph{arXiv preprint arXiv:2206.05514}, 2022.

\bibitem[Liang et~al.(2021)Liang, Cao, Sun, Zhang, Van~Gool, and Timofte]{swinir}
Jingyun Liang, Jiezhang Cao, Guolei Sun, Kai Zhang, Luc Van~Gool, and Radu Timofte.
\newblock Swinir: Image restoration using swin transformer.
\newblock In \emph{2021 IEEE/CVF International Conference on Computer Vision Workshops (ICCVW)}, pages 1833--1844, 2021.

\bibitem[Liu et~al.(2023)Liu, He, Gu, Kong, Qiao, and Dong]{degae}
Yihao Liu, Jingwen He, Jinjin Gu, Xiangtao Kong, Yu Qiao, and Chao Dong.
\newblock Degae: A new pretraining paradigm for low-level vision.
\newblock In \emph{Proceedings of the IEEE/CVF Conference on Computer Vision and Pattern Recognition}, pages 23292--23303, 2023.

\bibitem[Liu et~al.(2024)Liu, Ke, Liu, Zhao, and Lau]{diffplugin}
Yuhao Liu, Zhanghan Ke, Fang Liu, Nanxuan Zhao, and Rynson~WH Lau.
\newblock Diff-plugin: Revitalizing details for diffusion-based low-level tasks.
\newblock In \emph{Proceedings of the IEEE/CVF Conference on Computer Vision and Pattern Recognition}, pages 4197--4208, 2024.

\bibitem[Liu et~al.(2018)Liu, Jaw, Huang, and Hwang]{snow100k}
Yun-Fu Liu, Da-Wei Jaw, Shih-Chia Huang, and Jenq-Neng Hwang.
\newblock Desnownet: Context-aware deep network for snow removal.
\newblock \emph{IEEE Transactions on Image Processing}, 27\penalty0 (6):\penalty0 3064--3073, 2018.

\bibitem[Liu et~al.(2021)Liu, Lin, Cao, Hu, Wei, Zhang, Lin, and Guo]{swin}
Ze Liu, Yutong Lin, Yue Cao, Han Hu, Yixuan Wei, Zheng Zhang, Stephen Lin, and Baining Guo.
\newblock Swin transformer: Hierarchical vision transformer using shifted windows.
\newblock In \emph{Proceedings of the IEEE/CVF international conference on computer vision}, pages 10012--10022, 2021.

\bibitem[Luo et~al.(2023)Luo, Gustafsson, Zhao, Sj{\"o}lund, and Sch{\"o}n]{daclip}
Ziwei Luo, Fredrik~K Gustafsson, Zheng Zhao, Jens Sj{\"o}lund, and Thomas~B Sch{\"o}n.
\newblock Controlling vision-language models for universal image restoration.
\newblock \emph{arXiv preprint arXiv:2310.01018}, 3\penalty0 (8), 2023.

\bibitem[Ma et~al.(2023)Ma, Yang, Ju, Zhang, Liu, Wang, Zhang, and Wang]{diffusionseg}
Chaofan Ma, Yuhuan Yang, Chen Ju, Fei Zhang, Jinxiang Liu, Yu Wang, Ya Zhang, and Yanfeng Wang.
\newblock Diffusionseg: Adapting diffusion towards unsupervised object discovery.
\newblock \emph{arXiv preprint arXiv:2303.09813}, 2023.

\bibitem[Nah et~al.(2017)Nah, Hyun~Kim, and Mu~Lee]{gopro}
Seungjun Nah, Tae Hyun~Kim, and Kyoung Mu~Lee.
\newblock Deep multi-scale convolutional neural network for dynamic scene deblurring.
\newblock In \emph{Proceedings of the IEEE conference on computer vision and pattern recognition}, pages 3883--3891, 2017.

\bibitem[Potlapalli et~al.(2024)Potlapalli, Zamir, Khan, and Shahbaz~Khan]{promptir}
Vaishnav Potlapalli, Syed~Waqas Zamir, Salman~H Khan, and Fahad Shahbaz~Khan.
\newblock Promptir: Prompting for all-in-one image restoration.
\newblock \emph{Advances in Neural Information Processing Systems}, 36, 2024.

\bibitem[Quan et~al.(2021)Quan, Yu, Liang, and Yang]{rainds}
Ruijie Quan, Xin Yu, Yuanzhi Liang, and Yi Yang.
\newblock Removing raindrops and rain streaks in one go.
\newblock In \emph{Proceedings of the IEEE/CVF conference on computer vision and pattern recognition}, pages 9147--9156, 2021.

\bibitem[Radford et~al.(2021)Radford, Kim, Hallacy, Ramesh, Goh, Agarwal, Sastry, Askell, Mishkin, Clark, et~al.]{clip}
Alec Radford, Jong~Wook Kim, Chris Hallacy, Aditya Ramesh, Gabriel Goh, Sandhini Agarwal, Girish Sastry, Amanda Askell, Pamela Mishkin, Jack Clark, et~al.
\newblock Learning transferable visual models from natural language supervision.
\newblock In \emph{International conference on machine learning}, pages 8748--8763. PMLR, 2021.

\bibitem[Rajagopalan and Patel(2024)]{awracle}
Sudarshan Rajagopalan and Vishal~M Patel.
\newblock Awracle: All-weather image restoration using visual in-context learning.
\newblock \emph{arXiv preprint arXiv:2409.00263}, 2024.

\bibitem[Richter et~al.(2022)Richter, AlHaija, and Koltun]{epe}
Stephan~R Richter, Hassan~Abu AlHaija, and Vladlen Koltun.
\newblock Enhancing photorealism enhancement.
\newblock \emph{IEEE Transactions on Pattern Analysis and Machine Intelligence}, 45\penalty0 (2):\penalty0 1700--1715, 2022.

\bibitem[Rombach et~al.(2022)Rombach, Blattmann, Lorenz, Esser, and Ommer]{stablediff}
Robin Rombach, Andreas Blattmann, Dominik Lorenz, Patrick Esser, and Bj{\"o}rn Ommer.
\newblock High-resolution image synthesis with latent diffusion models.
\newblock In \emph{Proceedings of the IEEE/CVF conference on computer vision and pattern recognition}, pages 10684--10695, 2022.

\bibitem[Ruiz et~al.(2023)Ruiz, Li, Jampani, Pritch, Rubinstein, and Aberman]{dreambooth}
Nataniel Ruiz, Yuanzhen Li, Varun Jampani, Yael Pritch, Michael Rubinstein, and Kfir Aberman.
\newblock Dreambooth: Fine tuning text-to-image diffusion models for subject-driven generation.
\newblock In \emph{Proceedings of the IEEE/CVF conference on computer vision and pattern recognition}, pages 22500--22510, 2023.

\bibitem[Shen et~al.(2019)Shen, Wang, Lu, Shen, Ling, Xu, and Shao]{hide}
Ziyi Shen, Wenguan Wang, Xiankai Lu, Jianbing Shen, Haibin Ling, Tingfa Xu, and Ling Shao.
\newblock Human-aware motion deblurring.
\newblock In \emph{Proceedings of the IEEE/CVF international conference on computer vision}, pages 5572--5581, 2019.

\bibitem[Shipard et~al.(2023{\natexlab{a}})Shipard, Wiliem, Thanh, Xiang, and Fookes]{diffzeroshot1}
Jordan Shipard, Arnold Wiliem, Kien~Nguyen Thanh, Wei Xiang, and Clinton Fookes.
\newblock Diversity is definitely needed: Improving model-agnostic zero-shot classification via stable diffusion.
\newblock In \emph{Proceedings of the IEEE/CVF Conference on Computer Vision and Pattern Recognition}, pages 769--778, 2023{\natexlab{a}}.

\bibitem[Shipard et~al.(2023{\natexlab{b}})Shipard, Wiliem, Thanh, Xiang, and Fookes]{diffzeroshot2}
Jordan Shipard, Arnold Wiliem, Kien~Nguyen Thanh, Wei Xiang, and Clinton Fookes.
\newblock Boosting zero-shot classification with synthetic data diversity via stable diffusion.
\newblock \emph{arXiv preprint arXiv:2302.03298}, 3\penalty0 (5), 2023{\natexlab{b}}.

\bibitem[Toker et~al.(2024)Toker, Eisenberger, Cremers, and Leal-Taix{\'e}]{satsynth}
Aysim Toker, Marvin Eisenberger, Daniel Cremers, and Laura Leal-Taix{\'e}.
\newblock Satsynth: Augmenting image-mask pairs through diffusion models for aerial semantic segmentation.
\newblock In \emph{Proceedings of the IEEE/CVF Conference on Computer Vision and Pattern Recognition}, pages 27695--27705, 2024.

\bibitem[Valanarasu et~al.(2022)Valanarasu, Yasarla, and Patel]{transw}
J.~Jose Valanarasu, R. Yasarla, and V.~M. Patel.
\newblock Transweather: Transformer-based restoration of images degraded by adverse weather conditions.
\newblock In \emph{2022 IEEE/CVF Conference on Computer Vision and Pattern Recognition (CVPR)}, pages 2343--2353, 2022.

\bibitem[Wang et~al.(2019)Wang, Yang, Xu, Chen, Zhang, and Lau]{rain1}
Tianyu Wang, Xin Yang, Ke Xu, Shaozhe Chen, Qiang Zhang, and Rynson~W.H. Lau.
\newblock Spatial attentive single-image deraining with a high quality real rain dataset.
\newblock In \emph{2019 IEEE/CVF Conference on Computer Vision and Pattern Recognition (CVPR)}, pages 12262--12271, 2019.

\bibitem[Wei et~al.(2018)Wei, Wang, Yang, and Liu]{lolv1}
Chen Wei, Wenjing Wang, Wenhan Yang, and Jiaying Liu.
\newblock Deep retinex decomposition for low-light enhancement.
\newblock \emph{arXiv preprint arXiv:1808.04560}, 2018.

\bibitem[Wu et~al.(2023)Wu, Zhao, Shou, Zhou, and Shen]{diffumask}
Weijia Wu, Yuzhong Zhao, Mike~Zheng Shou, Hong Zhou, and Chunhua Shen.
\newblock Diffumask: Synthesizing images with pixel-level annotations for semantic segmentation using diffusion models.
\newblock In \emph{Proceedings of the IEEE/CVF International Conference on Computer Vision}, pages 1206--1217, 2023.

\bibitem[Xie et~al.(2022)Xie, Zhang, Cao, Lin, Bao, Yao, Dai, and Hu]{simmim}
Zhenda Xie, Zheng Zhang, Yue Cao, Yutong Lin, Jianmin Bao, Zhuliang Yao, Qi Dai, and Han Hu.
\newblock Simmim: A simple framework for masked image modeling.
\newblock In \emph{Proceedings of the IEEE/CVF conference on computer vision and pattern recognition}, pages 9653--9663, 2022.

\bibitem[Yang et~al.(2024)Yang, Kang, Huang, Xu, Feng, and Zhao]{depthanything}
Lihe Yang, Bingyi Kang, Zilong Huang, Xiaogang Xu, Jiashi Feng, and Hengshuang Zhao.
\newblock Depth anything: Unleashing the power of large-scale unlabeled data.
\newblock In \emph{Proceedings of the IEEE/CVF Conference on Computer Vision and Pattern Recognition}, pages 10371--10381, 2024.

\bibitem[Yasarla and Patel(2019)]{rainnew2}
Rajeev Yasarla and Vishal~M Patel.
\newblock Uncertainty guided multi-scale residual learning-using a cycle spinning cnn for single image de-raining.
\newblock In \emph{Proceedings of the IEEE/CVF conference on computer vision and pattern recognition}, pages 8405--8414, 2019.

\bibitem[Yasarla and Patel(2020)]{rainnew1}
Rajeev Yasarla and Vishal~M Patel.
\newblock Confidence measure guided single image de-raining.
\newblock \emph{IEEE Transactions on Image Processing}, 29:\penalty0 4544--4555, 2020.

\bibitem[Yin et~al.(2024)Yin, Gharbi, Zhang, Shechtman, Durand, Freeman, and Park]{dmd}
Tianwei Yin, Micha{\"e}l Gharbi, Richard Zhang, Eli Shechtman, Fredo Durand, William~T Freeman, and Taesung Park.
\newblock One-step diffusion with distribution matching distillation.
\newblock In \emph{Proceedings of the IEEE/CVF conference on computer vision and pattern recognition}, pages 6613--6623, 2024.

\bibitem[Yuan et~al.(2022)Yuan, Pinto, Davies, Gupta, and Torr]{diffclass3}
Jianhao Yuan, Francesco Pinto, Adam Davies, Aarushi Gupta, and Philip Torr.
\newblock Not just pretty pictures: Text-to-image generators enable interpretable interventions for robust representations.
\newblock \emph{arXiv preprint arXiv:2212.11237}, 3, 2022.

\bibitem[Zamir et~al.(2021)Zamir, Arora, Khan, Hayat, Khan, Yang, and Shao]{mprnet}
Syed~Waqas Zamir, Aditya Arora, Salman Khan, Munawar Hayat, Fahad~Shahbaz Khan, Ming-Hsuan Yang, and Ling Shao.
\newblock Multi-stage progressive image restoration.
\newblock In \emph{CVPR}, 2021.

\bibitem[Zamir et~al.(2022)Zamir, Arora, Khan, Hayat, Khan, and Yang]{restormer}
Syed~Waqas Zamir, Aditya Arora, Salman Khan, Munawar Hayat, Fahad~Shahbaz Khan, and Ming-Hsuan Yang.
\newblock Restormer: Efficient transformer for high-resolution image restoration.
\newblock In \emph{CVPR}, 2022.

\bibitem[Zhang et~al.(2020)Zhang, Sindagi, and Patel]{haze1}
He Zhang, Vishwanath Sindagi, and Vishal~M. Patel.
\newblock Joint transmission map estimation and dehazing using deep networks.
\newblock \emph{IEEE Transactions on Circuits and Systems for Video Technology}, 30\penalty0 (7):\penalty0 1975--1986, 2020.

\bibitem[Zhang et~al.(2021{\natexlab{a}})Zhang, Li, Yu, Luo, and Li]{snow1}
Kaihao Zhang, Rongqing Li, Yanjiang Yu, Wenhan Luo, and Changsheng Li.
\newblock Deep dense multi-scale network for snow removal using semantic and depth priors.
\newblock \emph{IEEE Transactions on Image Processing}, 30:\penalty0 7419--7431, 2021{\natexlab{a}}.

\bibitem[Zhang et~al.(2021{\natexlab{b}})Zhang, Dong, Pan, Zhu, Tai, Wang, Li, Huang, and Wang]{revide}
Xinyi Zhang, Hang Dong, Jinshan Pan, Chao Zhu, Ying Tai, Chengjie Wang, Jilin Li, Feiyue Huang, and Fei Wang.
\newblock Learning to restore hazy video: A new real-world dataset and a new method.
\newblock In \emph{CVPR}, pages 9239--9248, 2021{\natexlab{b}}.

\bibitem[Zhao et~al.(2023)Zhao, Ogawa, Chen, Yang, and Sekimoto]{labelfreedom}
Chenbo Zhao, Yoshiki Ogawa, Shenglong Chen, Zhehui Yang, and Yoshihide Sekimoto.
\newblock Label freedom: Stable diffusion for remote sensing image semantic segmentation data generation.
\newblock In \emph{2023 IEEE International Conference on Big Data (BigData)}, pages 1022--1030. IEEE, 2023.

\bibitem[Zheng et~al.(2024)Zheng, Wu, Yang, Zhang, Hu, and Zheng]{diffuir}
Dian Zheng, Xiao-Ming Wu, Shuzhou Yang, Jian Zhang, Jian-Fang Hu, and Wei-shi Zheng.
\newblock Selective hourglass mapping for universal image restoration based on diffusion model.
\newblock In \emph{Proceedings of the IEEE/CVF Conference on Computer Vision and Pattern Recognition}, 2024.

\end{thebibliography}
}



% % Supplementary References
% {
%     \small
%     \bibliographystyle{ieeenat_fullname}
%     \bibliography{supp}
% }

\end{document}